\documentclass[dvipsnames,format=sigconf]{acmart}

\AtBeginDocument{%
  }

\usepackage{graphicx}
\usepackage{amsmath}

\usepackage{booktabs} % For formal tables

\usepackage{colortbl}
\usepackage[ruled,commentsnumbered,linesnumbered,noend]{algorithm2e} % Algorithm 

\usepackage{amsmath,amsfonts,bm}

\def\eqref#1{equation~\ref{#1}}
\def\1{\bm{1}}

\DeclareMathAlphabet{\mathsfit}{\encodingdefault}{\sfdefault}{m}{sl}
\SetMathAlphabet{\mathsfit}{bold}{\encodingdefault}{\sfdefault}{bx}{n}

\newcommand{\xxnote}[3]{}
\ifx\hidenotes\undefined%
  \renewcommand{\xxnote}[3]{\color{#2}{(#1: #3)}}
\fi
\usepackage[normalem]{ulem}

\title{Multi-Objective Covariance Matrix Adaptation MAP-Annealing}

\author{Shihan Zhao}
\orcid{0009-0003-1344-5260}
\affiliation{%
  \institution{University of Southern California}
  \department{Thomas Lord Department of Computer Science}
  \city{Los Angeles} 
  \state{CA}
  \country{USA}
}
\email{shihanzh@usc.edu}

\author{Stefanos Nikolaidis}
\orcid{0000-0002-8617-3871}
\affiliation{%
  \institution{University of Southern California}
  \department{Thomas Lord Department of Computer Science}
  \city{Los Angeles} 
  \state{CA} 
  \country{USA}
}
\email{nikolaid@usc.edu}

\setcopyright{acmlicensed}
\copyrightyear{2025} 
\acmYear{2025} 
\setcopyright{cc} 
\setcctype{by} 
\acmConference[GECCO '25]{Genetic and Evolutionary Computation Conference}{July 14--18, 2025}{Malaga, Spain} 
\acmBooktitle{Genetic and Evolutionary Computation Conference (GECCO '25), July 14--18, 2025, Malaga, Spain}
\acmDOI{10.1145/3712256.3726463} 
\acmISBN{979-8-4007-1465-8/2025/07}

\begin{document}

% Abstract
\begin{abstract}
Quality-Diversity (QD) optimization is an emerging field that focuses on finding a set of behaviorally diverse and high-quality solutions. While the quality is typically defined w.r.t. a single objective function, recent work on Multi-Objective Quality-Diversity (MOQD) extends QD optimization to simultaneously optimize multiple objective functions. This opens up multi-objective applications for QD, such as generating a diverse set of game maps that maximize difficulty, realism, or other properties. Existing MOQD algorithms use non-adaptive methods such as mutation and crossover to search for non-dominated solutions and construct an archive of Pareto Sets (PS). However, recent work in QD has demonstrated enhanced performance through the use of covariance-based evolution strategies for adaptive solution search. We propose bringing this insight into the MOQD problem, and introduce MO-CMA-MAE, a new MOQD algorithm that leverages Covariance Matrix Adaptation-Evolution Strategies (CMA-ES) to optimize the hypervolume associated with every PS within the archive. We test MO-CMA-MAE on three MOQD domains, and for generating maps of a co-operative video game, showing significant improvements in performance.
% Attention: Abstract is limited to 200 words!
\end{abstract}




% Title
\maketitle

% Keywords
 % Keywords for indexing

\section{Introduction}

% Multi-Objective Quality Diversity (MOQD)~\citep{pierrot2022multi} extends the Quality Diversity (QD)~\citep{chatzilygeroudis2021quality} problem to consider multiple to-be-optimized objectives. Similar to QD, MOQD explores a behavior space defined by $b$ measure functions $m_1(.), \ldots, m_b(.)$, but instead of optimizing a single objective function $f(.)$ as in QD, MOQD optimizes $k > 1$ non-aligned objectives $f_1(.), \ldots, f_k(.)$. Since improving one objective often means compromising the others, MOQD aims to find a set of behaviorally diverse local Pareto Fronts (PF). The solutions contained within each PF must satisfy a specific combination of measure function constraints, and in the meantime, must achieve the best trade-off among all objectives.

Quality-Diversity (QD) is a growing field of optimization focused on finding a diverse archive of solutions that achieve high objective values. QD algorithms addressing this problem have been successfully applied across various scenarios, such as searching for control policies that exhibit distinct behaviors~\citep{cully2015robots, grillotti2022unsupervised}, generating niche testing scenarios~\citep{fontaine2022evaluating, bhatt2022deep}, and providing illustrative examples to probe and learn human preferences~\citep{ding2023quality, gallotta2023preference}. While the qualities of solutions are usually evaluated through a single objective function in the QD paradigm, recent work on Multi-Objective Quality-Diversity (MOQD)~\citep{pierrot2022multi, janmohamed2023improving} has extended QD to optimize multiple objectives. Instead of finding an archive of individual solutions, MOQD seeks to find a diverse archive of local Pareto Sets (PS). Each PS in the archive consists of solutions representing the optimal trade-offs among the objectives, while different local PSs capture solutions that exhibit different behaviors.

\begin{figure*}[!ht]
    \centering
    \includegraphics[width=0.9\linewidth]{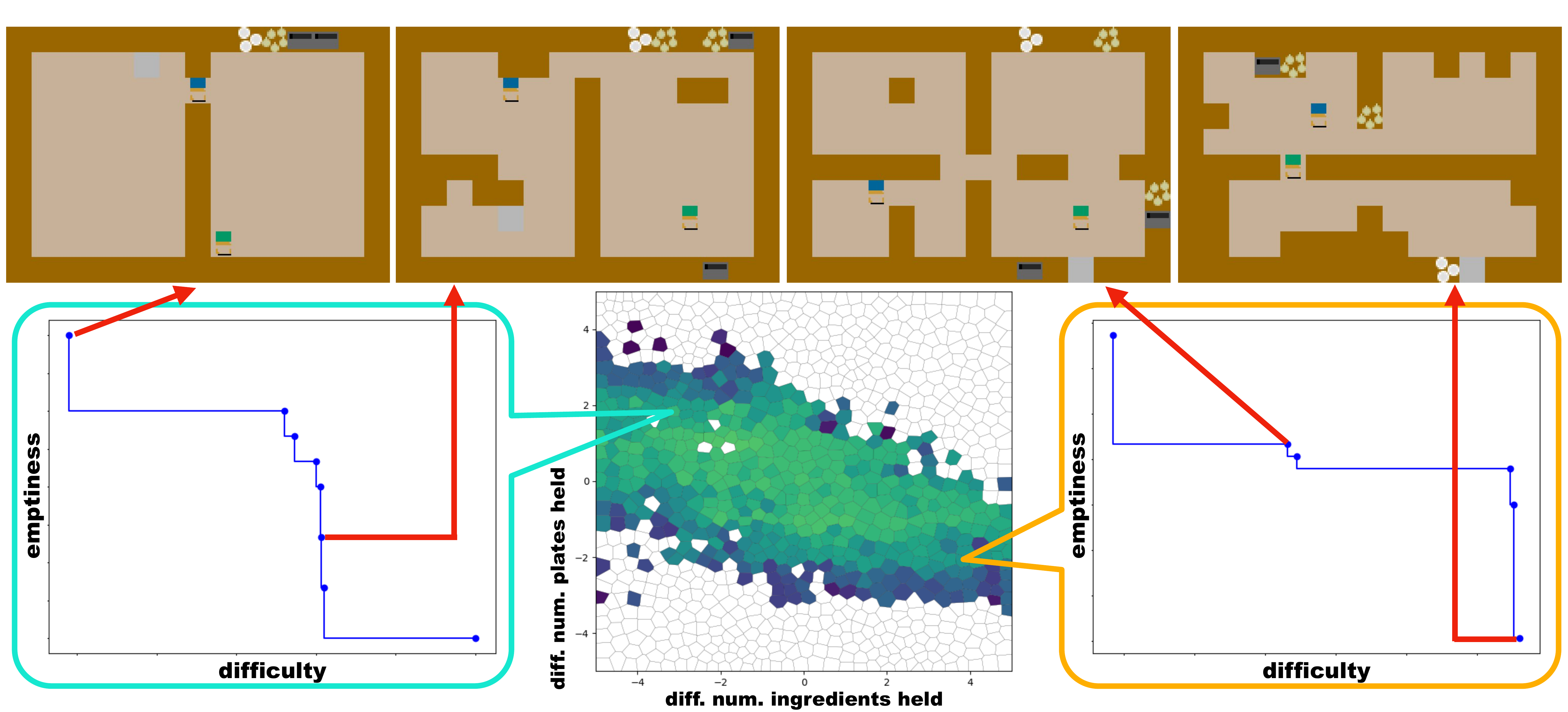}
    \caption{Game maps found by MO-CMA-MAE that induce diverse task divisions while achieving good trade-off between the \emph{difficulty} and \emph{emptiness} objectives. In this case, the two task divisions we consider are the difference between numbers of ingredients held by the two cooperating players, and difference between numbers of plates held. In the two upper-left maps, the blue player starts between the plate (represented by three white circles) and the delivery counter (represented by a gray tile), and thus is induced to specialize in delivering cooked meals on plates to the counter. The two upper-right maps induce the opposite, where the green player specializes in delivery.}
    \label{fig:overcooked}
\end{figure*}

The need to find behaviorally diverse PSs arises when a problem simultaneously features multiple objectives to be \emph{optimized}, and behavior-characterizing measure functions to be \emph{explored}. For example, when generating game level maps for a co-op maze game such as Overcooked~\citep{carroll2019utility}, we may want the generated maps to make it more difficult for players to achieve high scores, but in the meantime also cheaper to construct in the sense that more space is left empty. In addition to maximizing these two objectives, we may also want the maps to promote diversity in the playstyles they elicit from the cooperating players, ranging from specialized to even task divisions. We highlight the difference between \textbf{\textit{optimization}} and \textbf{\textit{exploration}}. When optimizing the objectives, we purposefully search for \emph{more} difficult and \emph{more} empty maps. However, when exploring the space spanned by measure functions, we want to cover \emph{all} induced playstyles, with no consistent preference on either even or uneven task divisions. In this particular application, we may use an MOQD algorithm to generate a behaviorally diverse archive of PSs, with each PS containing maps that balance difficulty and emptiness. This is illustrated in Figure~\ref{fig:overcooked}.

% Through an MOQD algorithm, we find an archive of behaviorally diverse PSs. Each PS consists of maps that optimize the trade-off between difficulty and emptiness, while collectively, the archive of PSs promotes diverse task divisions among the players. This is illustrated in Figure~\ref{fig:overcooked}.

To address the MOQD problem, prior work~\citep{pierrot2022multi, janmohamed2023improving} has proposed the Multi-Objective MAP-Elites (MOME) algorithm. Similar to the QD algorithm MAP-Elites~\citep{mouret2015illuminating}, MOME constructs a behaviorally-indexed archive by tessellating the measure space into discrete cells. Each solution is assigned into a cell according to its measure function outputs, and locally competes with other solutions assigned to this same cell. Unlike MAP-Elites, MOME does not maintain and update a single highest-objective elite within each cell, but an incumbent PS consisted of non-dominated solutions optimizing various trade-offs among all objectives. MOME shows promising results at finding a behaviorally diverse set of PSs on Rastrigin~\citep{hansen2021coco} and Brax~\citep{freeman2021brax} domains. However, MOME searchs for non-dominated solutions using mutation and crossover operations, which do not self-adapt according to the search progress. This limitation has been partly addressed in MOME-PGX~\citep{vassiliades2018discovering}, which employs policy gradient to accelerate optimization of solution objectives. Nonetheless, MOME-PGX does not explicitly use policy gradient to assist measure space exploration, and additionally assumes the problem to be a Markov Decision Process (MDP).

To overcome these limitations, we propose a new MOQD algorithm, Multi-Objective Covariance Matrix Adaptation MAP-Annealing (MO-CMA-MAE). MO-CMA-MAE utilizes the black-box optimizer Covariance Matrix Adaptation Evolution Strategy (CMA-ES)~\citep{hansen:cma16} to search for solutions that either explore the measure space by exhibiting an under-represented behavior, or optimizes a discovered behavior by achieving non-dominance within its corresponding PS. We highlight two key insights as contributions of this work:
\begin{enumerate}
    \item We recognize the hypervolume improvement of a solution over its archive cell as a metric that simultaneously motivates measure space exploration and multi-objective optimization, and optimize this metric with CMA-ES.
    \item To further incentivize CMA-ES to explore the measure space, we boost hypervolume improvements within under-explored archive cells by implementing a threshold acceptance mechanism inspired by the QD algorithm Covariance Matrix Adaptation MAP-Annealing (CMA-MAE)~\citep{fontaine2023covariance, zhaocovariance}.
\end{enumerate}

We benchmark MO-CMA-MAE on 4 test domains against MOME, NSGA-II~\citep{deb2002fast}, SMS-EMOA~\citep{beume2007sms}, and a modified COMO-CMA-ES~\citep{toure2019uncrowded}. MO-CMA-MAE outperforms all baselines in 2 of the domains. In the remaining 2 domains, MO-CMA-MAE performed on par with the strongest baseline MOME, while outperforming the other baselines.

\section{Background}
\subsection{Multi-Objective Optimization}
Multi-Objective Optimization (MOO) aims to find solutions that optimize various trade-offs among $k$ non-aligned objectives $f_i(\bm{x}); i \in \{1,2,\ldots, k\}$. Since multiple objectives are involved, when comparing two solutions $\bm{x}$ and $\bm{y}$, $\bm{y}$ only outperforms $\bm{x}$ when it is as good as $\bm{x}$ in all $k$ objectives, and strictly better in at least one objective (we assume maximization for MOO throughout this work). This is formalized as strict dominance:
\begin{equation}
    \bm{y} \succ \bm{x} \text{ iff } \forall i \in [k], f_i(\bm{y}) \geq f_i(\bm{x}) \text{ and } \exists j \in [k], f_j(\bm{y}) > f_j(\bm{x}) \addtocounter{equation}{1} \tag*{(1)}
\end{equation}

% \begin{equation}
%     y \succeq x \text{ iff } \forall i \in [k], f_i(\bm{y}) \geq f_i(\bm{x}) \addtocounter{equation}{1} \tag*{(weak dominance) \; (2)}
% \end{equation}

Dominance can also be used to compare sets, where $\bm{y} \succ \bm{x}$ iff $\forall \bm{x} \in X \; \exists \bm{y} \in Y \; s.t. \; \bm{y} \succ \bm{x}$, and between an individual and a set, where $\bm{y} \succ \bm{x} \text{ iff } \; \forall \bm{x} \in X, \bm{y} \succ \bm{x}$. A set of non-dominated solutions $P = \{\bm{x} \in \mathcal{X} \mid \not\exists \bm{y} \in \mathcal{X}, \bm{y} \succ \bm{x}\}$ from solution space $\mathcal{X}$ is known as the Pareto Set (PS). Pareto Front (PF) $F = \{f_{i \in [k]}(\bm{x}) \mid \bm{x} \in P\}$ is the image of PS in objective space. When the objective functions are continuous, there can be an infinite number of such solutions, so MOO algorithms often try to approximate PS and PF by finding a finite set of solutions whose objectives are evenly distributed along the PF. We will refer to PS and PF as their finite set approximations for the remainder of this work.

% Evolutionary Multi-Objective Algorithms (EMOA) are a family of MOO algorithms which use evolutionary algorithms as the optimizer. Generally speaking, evolutionary algorithms maintain a population of solutions, and iteratively improve the population by dropping solutions with lower fitnesses. This requires a way of ranking solutions, which can be implemented in various ways. Some EMOAs~\citep{zitzler2001spea2, igel2007covariance, nebro2009smpso} such as NSGA-II~\citep{deb2002fast} rank solutions by their extents of dominance on the rest of the population. Some other EMOAs~\citep{zhang2007moea, li2008multiobjective, li2011adaptive, li2014evolutionary} decompose the multi-objective optimization problem into local single-objective problems, each of which defines a ranking for a portion of the solutions. 

% Hypervolume is strictly monotonic\citep{berghammer2010set} and its ranking is invariant to objective scales~\citep{guerreiro2021hypervolume}.

\subsection{The Hypervolume Indicator}
Quality indicators~\citep{trautmann2013r2, bringmann2014two, ishibuchi2015modified} are set functions that evaluate the optimality, and sometimes diversity, of a PF. Although quality indicators are often used as performance metric when benchmarking MOO algorithms, they may also be optimized directly using blackbox evolutionary algorithms. This combination led to the development of several indicator-based multi-objective evolutionary algorithms~\citep{zitzler2004indicator, diaz2013ranking, menchaca2015gd, tian2017indicator}. One such quality indicator is the hypervolume indicator, defined as the objective-space volume bounded between the evaluated PF and a reference point $r$, usually chosen to be slightly worse than the worst possible value along every objective. Formally, hypervolume may be defined as a union of polytopes~\citep{zitzler1999multiobjective}:
\begin{equation}
    \mbox{HV}_r(F) = \Lambda \left( \bigcup_{{\bf f}(\bm{x}) \in F} [{\bf f}(\bm{x}), r] \right)
\end{equation}

Where $\Lambda$ is the Lebesgue measure, and $[{\bf f}(\bm{x}), r]$ denotes the polytope in objective space delimited above by ${\bf f}(\bm{x})$ and below by $r$. MOEAs often optimize hypervolume incrementally~\citep{beume2007sms, igel2007covariance, jiang2014simple}, and need to evaluate each individual solution rather than the entire PF. In this case, an incremental case of hypervolume such as hypervolume improvements (HVI)~\citep{toure2019uncrowded} may be optimized:
\begin{equation}
    \label{eq:hvi}
    \mbox{HVI}_r({\bf f}(\bm{x}), F) = \mbox{HV}_r(F \cup \{{\bf f}(\bm{x})\}) - \mbox{HV}_r(F)
\end{equation}

\subsection{CMA-ES}
Covariance Matrix Adaptation Evolution Strategy (CMA-ES)~\citep{hansen:cma16} is among the most competitive blackbox optimizers for single-objective functions. CMA-ES samples likely high-objective solutions from a multivariate normal distribution $\mathcal{N}(\bm{\bar{x}}, \mathcal{C})$ with mean $\bm{\bar{x}}$ and covariance matrix $\mathcal{C}$, and uses their evaluation results to adapt $\mathcal{N}(\bm{\bar{x}}, \mathcal{C})$ to generate yet better solutions.

Although CMA-ES typically optimizes a single-objective function, it can be adapted for multi-objective optimization by optimizing a scalar metric that is consistent with Pareto dominance. Dominance-based ranking is one such metric, and some MOEAs such as MO-CMA-ES use CMA-ES to optimize dominance-based ranking while relying on a quality indicator to resolve ambiguities among mutually non-dominated solutions~\citep{igel2007covariance, voss2009recombination, voss2010improved}. Alternatively, some other MOEAs directly optimize a quality indicator with CMA-ES~\citep{igel2007covariance, voss2009recombination, voss2010improved}.

% MO-CMA-ES~\citep{igel2007covariance} and COMO-CMA-ES~\citep{toure2019uncrowded, gharafi2023multiobjective} are two prominent MOO algorithms that use CMA-ES to optimize hypervolume indicator.
% and defines a ranking which is invariant to objective scales~\citep{guerreiro2021hypervolume}. 

\subsection{Multi-Objective Quality-Diversity}
Quality-Diversity (QD)~\citep{mouret2015illuminating, vassiliades2016scaling, nilsson2021policy, fontaine2023covariance, batra2023proximal} is a family of algorithms aiming to find a set of solutions that are high-quality w.r.t. an objective function, and diverse w.r.t. some measure functions. Given a single objective function $f: \mathbb{R}^n \rightarrow \mathbb{R}$ and $M$ measure functions $m_{i \in [M]}: \mathbb{R}^n \rightarrow \mathbb{R}$, QD discretizes the measure space spanned by $m_{i \in [m]}$, and fills each cell in this discretization with a highest-$f$ solution whose measure values fall within this cell. This creates a measure-indexed archive that stores a single solution at every combination of discretized measure values. QD algorithms often leverage multiple variation operators in parallel to generate solutions to populate this archive. The variation operators are sometimes referred to as \emph{emitters}, and have varying implementations for different QD algorithms. For example, MAP-Elites~\citep{mouret2015illuminating} uses a mix of mutation and crossover emitters, whereas CMA-ME~\citep{fontaine2020covariance} uses CMA-ES emitters.

While QD features a single objective function $f$, Multi-Objective Quality-Diversity (MOQD)~\citep{pierrot2022multi, janmohamed2023improving} extends the QD framework to consider multiple non-aligned objective functions $f_1, f_2, \ldots, f_k$. Since multiple objective functions are involved, MOQD aims to find a local PS within every cell instead of a single solution. Formally, MOQD aims to optimize the sum of some quality indicator across all cells. This objective function is defined as the MOQD-score, and in the following definition, we use hypervolume for the quality indicator:
\begin{equation}
    \label{eq:moqd_score}
    \sum_{e=1}^{|A|} \mbox{HV}(F_e)
\end{equation}

where $|A|$ is the number of archive cells, and $F_e$ is the PF corresponding to the PS within cell $e$.

\section{Prior Works}

\subsection{CMA-MAE}
Covariance Matrix Adaptation MAP-Elites (CMA-ME)~\citep{fontaine2020covariance} and Covariance Matrix Adaptation MAP-Annealing (CMA-MAE)~\citep{fontaine2023covariance, zhaocovariance} are two QD algorithms that leverage CMA-ES to adaptively search for candidate solutions that improve the QD archive. When a candidate solution is inserted into an archive cell that matches its measure values, the difference between the candidate's and the cell's objective values is taken as the \textit{improvement score} of this candidate.

In CMA-ME, \textit{improvement score} is computed as $\Delta_A(\bm{x}) = f(\bm{x}) - f_A$, where $f(\bm{x})$ is the candidate's objective value, and $f_A$ is the objective of the predecessor solution within this candidate's cell. Candidates are ranked by their improvement scores, and CMA-ES learns from the ranking to generate candidate solutions that maximize improvement, either through having high objective values, or through having measure values that match them against low-objective predecessors. However, as noted in subsequent works~\citep{tjanaka2022approximating, fontaine2023covariance}, CMA-ME tends to prematurely abandon improving each cell, since a cell's $f_A$ increases quickly with every inserted solution, and the improvement scores for subsequent candidates deplete as $f(\bm{x}) - f_A \rightarrow 0$.

Instead of computing the improvement score as $\Delta_A(\bm{x}) = f(\bm{x}) - f_A$, CMA-MAE maintains a threshold value $t_e$ within each cell, and computes $\Delta_A(\bm{x}) = f(\bm{x}) - t_e$. If $f(\bm{x}) > t_e$ and the solution is inserted, $t_e$ is softly updated with $t_e \gets (1 - \alpha) t_e + \alpha f(\bm{x})$, where $\alpha \in (0, 1)$ is a learning rate hyperparameter. Due to the soft update, $t_e$ gradually approximates $f(\bm{x})$ as cell $e$ is visited with $f(\bm{x})$ more times. This delays the depletion of improvement score $f(\bm{x}) - t_e$ so that CMA-ES has more opportunities to improve each cell, and crucially, introduces measure space density descent as $t_e$ tends to be smaller and $f(\bm{x}) - t_e$ larger for cells that have been visited fewer times.

\subsection{MOME}
Multi-Objective MAP-Elites (MOME)~\citep{pierrot2022multi} proposes the first MOQD algorithm by modifying MAP-Elites~\citep{mouret2015illuminating} with an MOQD archive. Instead of a single solution, the MOQD archive maintains an incumbent PS (and the corresponding PF) within every cell. Similar to MAP-Elites, MOME generates candidate solutions with mutation and crossover-based operators, and assigns them to archive cells according to their measure values. A solution is inserted either when its assigned cell is empty, or when it is non-dominated by the incumbent PS within its cell. If it is inserted, the solutions it dominates and their objectives are dropped from the incumbent PS and PF respectively, so that the incumbent PS remains a (locally) non-dominated set. For memory-constrained applications, MOME implements an upper bound on the maximum set size within every cell, and drops solutions randomly from cells exceeding the size limit.

Subsequent work~\citep{janmohamed2023improving} further augments MOME with crowding-based selection and replacement. When generating candidate solutions with crossover or mutation, MOME samples parent solutions with probabilities proportional to their crowding distances. Additionally, when a cell exceeds the size limit, the solution with the smallest crowding distance is dropped first. Through these two mechanisms, MOME spreads out the solutions evenly in objective space. \citet{janmohamed2023improving} also proposes MOME-PGX, which learns a policy-gradient for every objective to accelerate its optimization. However, MOME-PGX assumes an MDP setting, which is more restrictive than the episodic MOQD problems we focus on in this work. Therefore, we only benchmark against the augmented version of MOME.

\section{Multi-Objective Covariance Matrix Adaptation MAP-Annealing}
\label{sec:algorithm}

Although MOME shows promising results at finding a diverse archive of PSs, it employs non-adaptive mutation and crossover-based solution search, which does not explicitly bias towards solutions that achieve non-dominance or exhibit new behaviors. To incorporate adaptive search into MOQD optimization, we propose using CMA-ES to explicitly search for solutions that optimize the MOQD objective.

\subsection{Using CMA-ES to search for large HVI}
\label{sec:rank_by_HVI}
In our proposed approach, we use CMA-ES to iteratively generate batches of candidate solutions to be inserted into the MOQD archive, and adapt CMA-ES according to some function over the insertion outcomes. Ideally, this should be a scalar-valued function that assigns higher values to solutions that either explore the measure space by discovering new archive cells, or locally optimize objectives by expanding the PF within a discovered cell. Both requirements are met by the hypervolume improvement $\mbox{HVI}_r({\bf f}(\bm{x}), T_e)$, where $\bm{x}$ is a candidate solution, $e$ is the archive cell assigned to $\bm{x}$ according to its measure values, and $T_e$ is the existing front within cell $e$. When cell $e$ is newly discovered, the first solution that discovers $e$ will have its hypervolume improvement computed relative to $T_e = \emptyset$, resulting in a large hypervolume improvement equal to the product of all its objectives (after subtracting the reference point $r$). When cell $e$ has been visited during previous iterations, the hypervolume improvement over $T_e$ is positive only for solutions non-dominated by $T_e$, and increases further with the margin of non-dominance.

Intuitively, hypervolume improvement is comparable to the concept of \textit{improvement score} from single-objective QD~\citep{fontaine2020covariance}. By adapting CMA-ES to generate solutions with larger hypervolume improvements and inserting them into the archive, MO-CMA-MAE incrementally optimizes the hypervolumes of every cell, thereby optimizing the MOQD score (Eq.~\ref{eq:moqd_score}) of the entire archive. As an additional benefit, hypervolume is currently the only unary indicator that is strictly Pareto-compliant, which guarantees its optimization does not lead to the deterioration of the corresponding PF~\citep{berghammer2010set, shang2020survey}. This provides consistent gradient information for CMA-ES to exploit, and prevents deteriorative cycle in which CMA-ES re-discovers a local PF that is dominated by its predecessors.

\subsection{Threshold Accepting Mechanism}
\label{sec:threshold_soft}
\noindent\textbf{Motivation.} By using CMA-ES to search for solutions with large hypervolume improvements within their assigned cells, we incentivize it to either explore the measure space for new cells or improve discovered cells. This is reminiscent of CMA-ES-driven QD algorithms such as CMA-ME~\citep{fontaine2020covariance}. However, measure space exploration and cell improvement compete for CMA-ES's attention, and while we desire CMA-ES to balance between these two aspects, we observe that it tends to overly focus on cell improvement in MOQD. This is because as the number of objectives, and thereby the objective space volume, increases, candidate solutions are more likely to be trivially non-dominated~\citep{purshouse2007evolutionary}. Consequently, compared to single-objective QD, multi-objective QD allows CMA-ES to more easily obtain frequent but small HVI rewards within discovered cells, thereby keeping its attention on cell improvement and away from measure space exploration. In order to encourage CMA-ES to explore the measure space in MOQD, we need to give it larger and more frequent rewards for visiting under-explored cells.

\noindent\textbf{Threshold Front with Discounted Objectives.} 
% This is very high-level; we need to explain how the threshold front approximates the real front as more solutions are inserted, and how it relates to HVI annealing.
To boost hypervolume improvements within under-explored cells, MO-CMA-MAE maintains a \textit{threshold front} $T_e$ in each cell $e$ that ``lags behind'' the local PF $F_e$, and computes the hypervolume improvement $\mbox{HVI}_r({\bf f}(\bm{x}), T_e)$ relative to $T_e$ instead of $F_e$. When a cell $e$ is newly discovered, MO-CMA-MAE keeps the gap between $T_e$ and $F_e$ initially large, which massively boosts hypervolume improvements for the early solutions assigned to $e$. As more solutions are inserted into $e$ and it becomes better-explored, MO-CMA-MAE gradually closes the gap between $T_e$ and $F_e$, thereby mitigating the boost and encouraging CMA-ES to shift its attention towards other under-explored cells, where the boost remains high. We note that MO-CMA-MAE inserts a solution $\bm{x}$ into its assigned cell $e$ only when it is non-dominated by $T_e$, i.e., when $\mbox{HVI}_r({\bf f}(\bm{x}), T_e) > 0$. The gradually rising $T_e$ may thus be viewed as implementing a \textit{threshold accepting} mechanism, similar to that employed in the single-objective QD algorithm CMA-MAE~\citep{fontaine2023covariance}. By gradually tightening each cell's acceptance criterion via its rising threshold front, MO-CMA-MAE effectively \textit{anneals} the hypervolume within each cell.

\begin{figure}[!h]
    \centering
    \includegraphics[width=0.8\columnwidth]{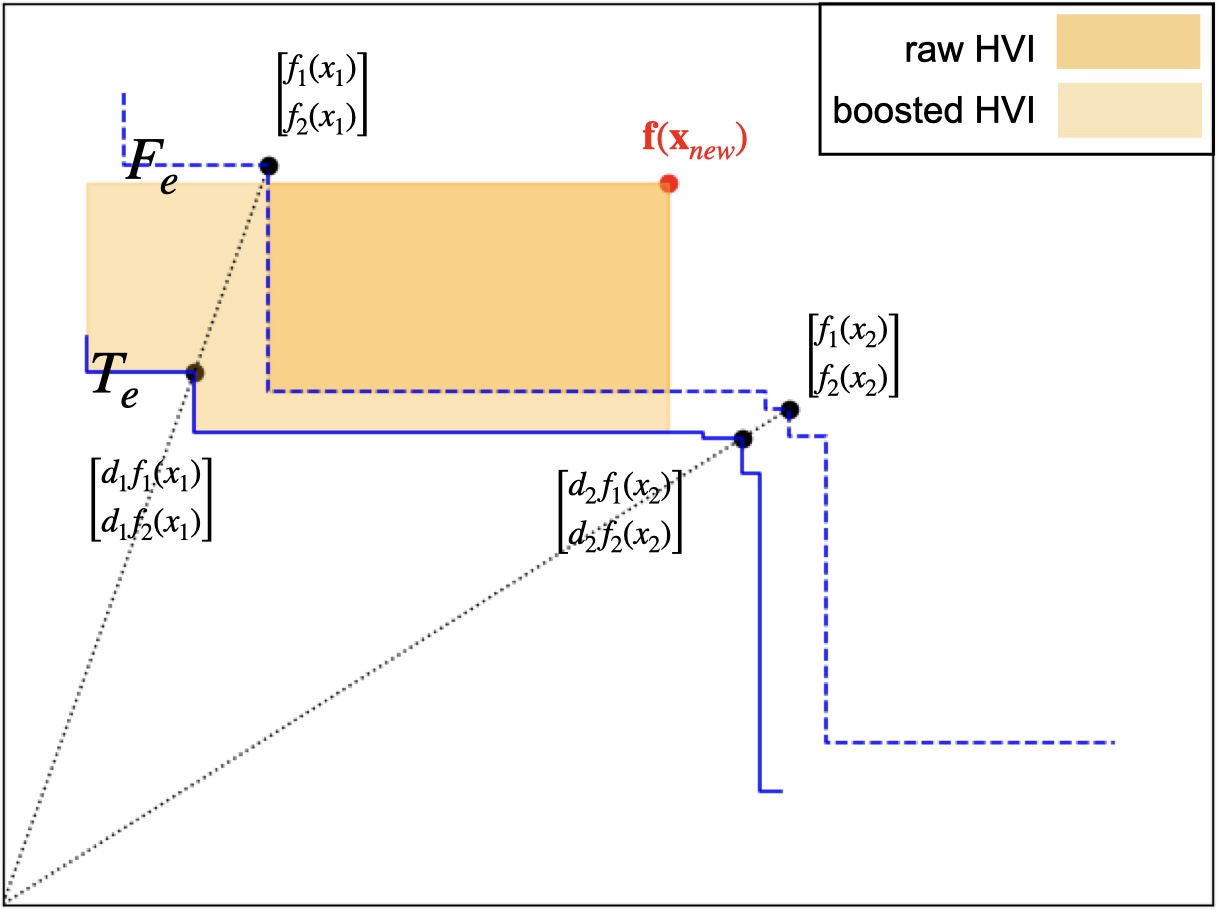}
    \caption{The gap between $T_e$ and $F_e$ boosts HVI for a new solution, and should be narrower at well-explored trade-offs.}
    \label{fig:tpf}
\end{figure}

To introduce the desired lag, we construct $T_e$ for each cell $e$ by iteratively inserting discounted objectives. When a solution $\bm{x}$ is assigned to $e$ and qualifies for insertion with $\mbox{HVI}_r({\bf f}(\bm{x}), T_e) > 0$, we multiply all its objectives with some discount factor $d \in [0,1]$ and insert $d{\bf f}(\bm{x})$ into $T_e$. Since the same $d$ is applied to all objectives, $d{\bf f}(\bm{x})$ retains the inter-objective ratios of ${\bf f}(\bm{x})$, and thus explores the same trade-off. Furthermore, at this trade-off, we may control through $d$ the local gap between $F_e$ and $T_e$ after inserting $d{\bf f}(\bm{x})$, where the local gap remains unchanged if $d$ is too small such that $d{\bf f}(\bm{x}) \prec T_e$, and closes completely if $d = 1$ (assuming locally Pareto-optimal $\bm{x}$). We need the gap to close gradually, so we need to search for a $d$ between these two extremes. Before we describe this search, we note that the value of $d_i$ may vary for each $\bm{x}_i$ depending on its objective values ${\bf f}(\bm{x}_i)$ and the current $T_e$. While this requires us to perform the discount factor search on every $\bm{x}_i$ to be inserted, having fine control over every $d_i{\bf f}(\bm{x}_i)$ also allows us to localize the gap between $T_e$ and $F_e$ at each trade-off, with narrower gaps at well-explored trade-offs. We illustrate in Figure~\ref{fig:tpf} how the gap between $T_e$ and $F_e$ boosts HVI for a new solution, as well as how the gap should ideally be localized.

\begin{figure*}[!ht]
    \centering
    \includegraphics[width=0.8\linewidth]{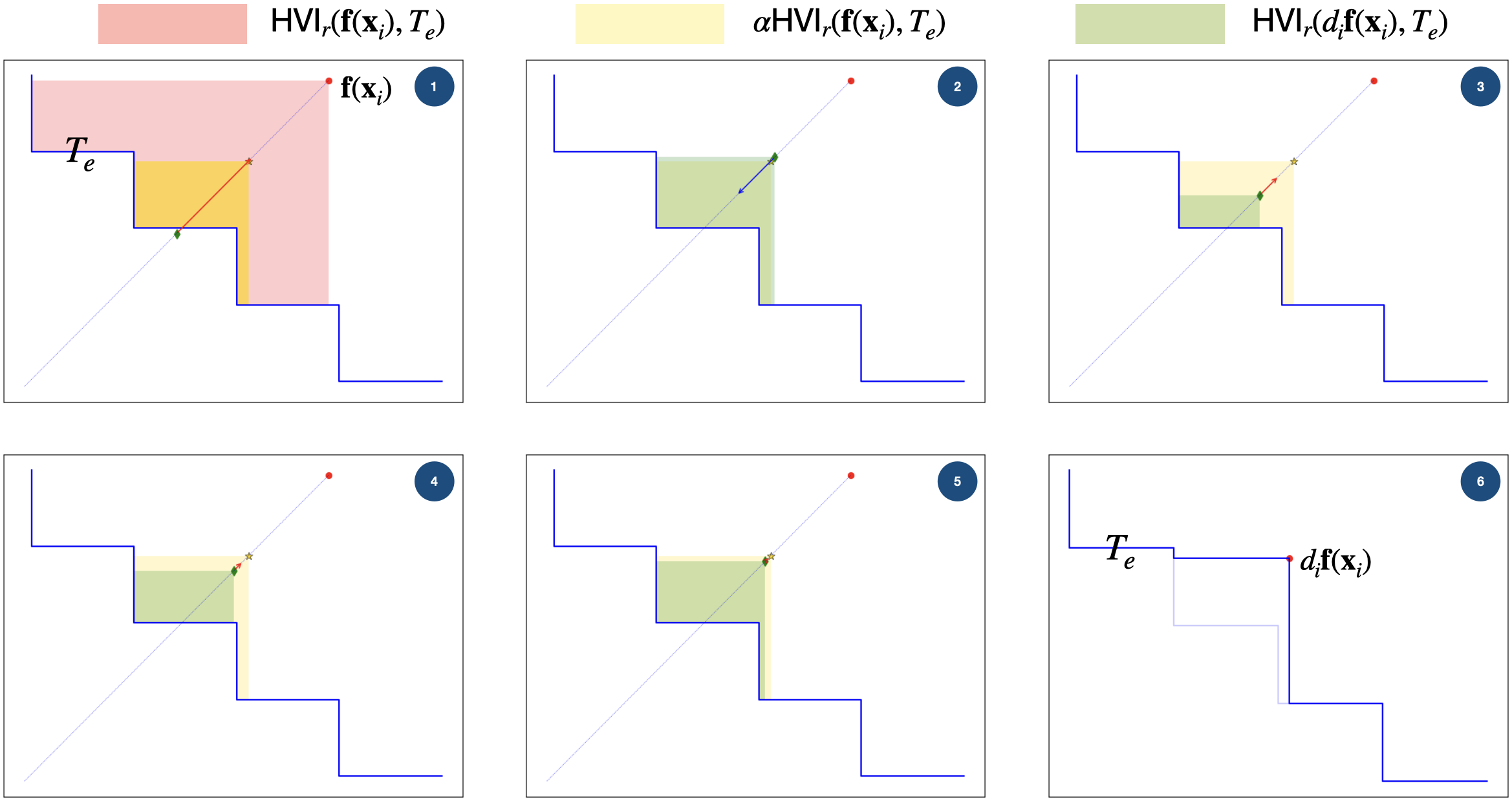}
    \caption{When inserting a new solution with objectives ${\bf f}({\bf x}_i)$ to its assigned threshold front $T_e$, MO-CMA-MAE searches for a discount factor $d_i$ satisfying $\mbox{HVI}_r(d_i{\bf f}({\bf x}_i), T_e) \approx \alpha\mbox{HVI}_r({\bf f}({\bf x}_i), T_e)$.}
    \label{fig:bisection}
\end{figure*}

\noindent\textbf{Discount Factor Search.} When searching the discount factor $d_i$ for a solution $\bm{x}_i$ assigned to cell $e$, our goal is to ensure that inserting $d_i{\bf f}(\bm{x}_i)$ to $T_e$ closes its gap to $F_e$ by a fraction. To quantify this goal, we use the hypervolume improvement of the discounted objectives over the pre-insertion threshold front, i.e., $\mbox{HVI}^{disc}(d_i) = \mbox{HVI}_r(d_i{\bf f}(\bm{x}_i), T_e)$, as a heuristic for the gap between $T_e$ and $F_e$ closed by inserting $d_i{\bf f}(\bm{x}_i)$. We observe that $\mbox{HVI}^{disc}(d_i)$ is a strictly monotonic function over $d_i \in [0,1]$, and is upper-bounded by the hypervolume improvement of the raw, undiscounted objectives $\mbox{HVI}^{disc}(1) = \mbox{HVI}_r({\bf f}(\bm{x}_i), T_e)$. This allows us to perform bisection search over $d_i \in [0,1]$, terminating when $\mbox{HVI}^{disc}(d_i) \approx \alpha\mbox{HVI}^{disc}(1) = \alpha\mbox{HVI}_r({\bf f}(\bm{x}_i), T_e)$. In other words, we find a $d_i$ that closes a fraction $\alpha$ of the maximum closable gap. We refer to the fraction hyperparameter $\alpha \in (0,1)$ as MO-CMA-MAE's \textit{learning rate}, because it can be interpreted as the learning rate of each cell's hypervolume. Figure~\ref{fig:bisection} illustrates an example run of this bisection search, in which the target HVI (yellow) is $\alpha = 0.2$ of the raw HVI (red), and the search pointer (green) approximates the target after a few iterations. We provide further details on bisection search implementation in Appendix~\ref{sec:bisec}.

\begin{algorithm}[h!]
\SetAlgoLined
\caption{MO-CMA-MAE}
\label{alg:mo_cma_mae}
\SetKwInOut{Input}{input}
\SetKwInOut{Result}{result}
\SetKwProg{MOCMAMAE}{MO-CMA-MAE}{}{}
\DontPrintSemicolon
\MOCMAMAE{$(N, \psi, \lambda, \bm{x}_0, \sigma_0$, $\alpha$, $\epsilon)$}
{

\Input{A desired number of iterations $N$, the number of emitters $\psi$, a branching population size $\lambda$, an initial solution $\bm{x}_0$, an initial step size $\sigma_0$, a learning rate $\alpha$, and a bisection search error tolerance $\epsilon$.}
\Result{An archive $A$ of Pareto Sets. Each cell $e$ within $A$ contains a Pareto Set $P_e$ whose solutions are non-dominated by other solutions from cell $e$.}

\BlankLine
Initialize archive $A$ where $\forall e \in A, \; P_e \gets \emptyset, \; T_e \gets \emptyset$.

Initialize $\psi$ CMA-ES emitters, each with initial mean $\bm{\bar{x}} \leftarrow \bm{x}_0$, initial covariance matrix $\mathcal{C} \leftarrow \sigma_0\mathbf{I}$, and other internal parameters $\bm{p}$.

\For{$iter\leftarrow 1$ \KwTo $N$}{
    \For{$i\leftarrow 1$ \KwTo $\lambda$}{
        \label{mocmamae:startforloop} 
        $\bm{x}_i \sim \mathcal{N}(\bm{\bar{x}},\mathcal{C})$\; \label{mocmamae:sample} 
        
        ${\bf f}(\bm{x}_i), \bm{m}(\bm{x}_i) \gets \mbox{evaluate}(\bm{x}_i)$\; \label{mocmamae:evaluate}
        
        $e \gets \mbox{calculate\_cell}(A, \bm{m}(\bm{x}_i))$\; \label{mocmamae:calc_cell}
        
        \BlankLine
        \tcc{Add $\bm{x}_i$ to $P_e$ if ${\bf f}(\bm{x}_i)$ is non-dominated by $T_e$; then search for $d_i$ and add $d_i{\bf f}(\bm{x}_i)$ to $T_e$}
        $\Phi_i \gets \mbox{HVI}_r({\bf f}(\bm{x}_i), T_e)$\; \label{mocmamae:calc_hvi}
        \If{$\Phi_i > 0$}{
            $P_e \gets \{\bm{x}_i\} \cup \{\bm{x} \in P_e \mid \bm{x} \not\prec \bm{x}_i\}$\; \label{mocmamae:add_to_ps}

            $d_i \gets \mbox{bisect}({\bf f}(\bm{x}_i), T_e, \alpha, \epsilon)$\;
            
            $T_e \gets \{d_i{\bf f}(\bm{x}_i)\} \cup \{{\bf f} \in T_e \mid {\bf f} \not\prec d_i{\bf f}(\bm{x}_i)\}$\; \label{mocmamae:add_to_tpf}
            $\text{(Optional) } \mbox{downsize}(P_e, T_e)$\; \label{mocmamae:downsize}
         }

    \BlankLine
    \tcc{Adapt CMA-ES~\citep{pyribs} according to $\Phi_i$}
    rank $\bm{x}_i$ by $\Phi_i$ \; \label{mocmamae:ranking}
    adapt CMA-ES parameters $\bm{\bar{x}},\mathcal{C},\bm{p}$\; \label{mocmamae:adaptation}
    \If{\mbox{CMA-ES restart triggered}}{
        $\bm{x}_0 \gets \mbox{sample\_elite}(\{P_e, \forall e \in A\})$\;
        Restart CMA-ES with $\mathcal{C} \leftarrow \sigma_0 I$ and $\bm{\bar{x}} \leftarrow \bm{x}_0$.
    }
}
}
}

\end{algorithm}

\subsection{Crowding-Based Downsizing}
\label{sec:downsize}
In real-world applications, it is generally desirable to control memory usage of an algorithm below some limit. For this purpose, we note that our threshold front is compatible with the crowding-distance-based downsizing mechanism introduced in~\citep{janmohamed2023improving}. If the number of solutions inserted into $P_e$ exceeds a certain size limit, a solution that has the smallest objective-space crowding distance is dropped. This approach prioritizes retaining solutions from less crowded regions, ensuring that the threshold front remains diverse. One additional complication introduced by downsizing is that $T_e$ retracts after each drop, and reopens already-covered objective space regions. This might lead to a cycle in which CMA-ES repeatedly inserts the same solution into a saturated $T_e$, and the solution repeatedly gets dropped. To detect and escape from this cycle, we implement a \textit{cycle} restart rule for CMA-ES. This restart rule counts the number of times each cell within the archive is visited, and restarts CMA-ES when a solution it generates falls within a cell that has been visited more than $10$ times the average number of visits across all cells. We refer to the version of MO-CMA-MAE constrained by a $T_e$ size limit as the \textit{static archive version}, and the version without a size constraint on $T_e$ as the \textit{dynamic archive version}. We primarily benchmark the \textit{static archive version} of MO-CMA-MAE to ensure it adheres to the same memory budget as other baselines, but we include an ablation experiment in Appendix~\ref{sec:staticvdynamic} to show that the two versions have comparable performances. Collectively, these components constitute the full MO-CMA-MAE algorithm, as outlined in Algorithm~\ref{alg:mo_cma_mae}.

%   \label{mocmamae:endforloop}
% rank $\bm{\theta}_i$ by $\Phi_i$ \; \label{mocmamae:ranking}

% adapt CMA-ES parameters $\bm{\theta},\Sigma,\bm{p}$\; \label{mocmamae:adaptation}

% \If{\mbox{CMA-ES converges}}{
% $\bm{\theta_0} \gets \mbox{sample\_elite}(A_{T_e})$

% Restart CMA-ES with $\Sigma \leftarrow \sigma_0 I$ and $\bm{\theta} \leftarrow \bm{\theta_0}$.
% }
% }

% }

\section{Experiments and Results}

We evaluate the performance of MO-CMA-MAE across the \textit{sphere}, \textit{rastrigin}, \textit{arm}, and \textit{overcooked} domains. Among these, \textit{sphere} and \textit{rastrigin} are numerical domains from prior work~\citep{fontaine2020covariance, hansen2021coco} modified to evaluate two objectives, each with a distinct shifted extremum; \textit{arm} is likewise modified from prior work~\citep{cully2017quality} to evaluate two objectives each based on the variance of half the joints; \textit{overcooked} is an adapted scenario generation task~\citep{carroll2019utility, fontaine2021importance} that aims to generate a set of game maps that are challenging, feature few wall tiles, and induce diverse player strategies. We provide more details on domain implementations in Appendix~\ref{sec:domains}.

\subsection{Baselines}

To benchmark MO-CMA-MAE, we choose the latest version of MOME~\citep{janmohamed2023improving}, NSGA-II~\citep{deb2002fast}, SMS-EMOA~\citep{beume2007sms}, and a modified COMO-CMA-ES~\citep{toure2019uncrowded} as our baselines. We provide details on how we modified COMO-CMA-ES to facilitate its comparison with the other baselines in Appendix~\ref{sec:comocmaes}. We outline some other baseline implementation choices we made in the interest of fair comparison below.

\noindent\textbf{Controlling Population Sizes.} We control computational resource usage by restricting all baselines to memorize at most $10000$ solutions. In the case of MO-CMA-MAE and MOME, this constraint is enforced by initializing their MOQD archives to have $1000$ cells, and keeping the number of solutions within each cell below $10$ through downsizing (described in Sec~\ref{sec:downsize}). In the case of NSGA-II, SMS-EMOA, and COMO-CMA-ES, this simply means setting their population size limit to $10000$. To enable MOQD-score (Eq.~\ref{eq:moqd_score}) computation for the non-QD baselines, we assign a passive MOQD archive with $1000$ cells. The passive archive has no size constraints and does not interact with the primary algorithm, other than storing all non-dominated solutions discovered during its execution.

\noindent\textbf{Implementation Details.} We implement MO-CMA-MAE from the pyribs library~\citep{pyribs}. The MOME and NSGA-II baselines are adapted from the QDax~\citep{chalumeau2023qdax} library. The SMS-EMOA and COMO-CMA-ES baselines are adapted from the pymoo~\citep{pymoo} and moarchiving~\citep{hansenmoarchiving} libraries, respectively. All MOQD archives are tesselated with the CVT~\citep{vassiliades2016scaling} approach after $50000$ samples. For the toy domain experiments, we use $5$ emitters each with batch size $36$, and run $5000$ iterations for each algorithm. This sums up to $900000$ evaluations for each algorithm. For overcooked level generation, we use $1$ emitter with batch size $16$, and run $5000$ iterations for each algorithm. Each solution is aggregated over $4$ evaluations. This sums up to $320000$ evaluations for each algorithm. We choose $\epsilon = 1 \times 10^{-3}$ and $\alpha = 0.1$ for MO-CMA-MAE when comparing against baselines. Our choices for the remaining hyperparameters are listed in Appendix~\ref{sec:hyperparams}.

\begin{table*}[!ht]
\caption{Mean MOQD-score and Coverage values achieved at the end of 5000 iterations.}
\label{tab:results_new_domains}\centering\resizebox{0.9\linewidth}{!}{\begin{tabular}{lrrrrrrrr}
\toprule
 & \multicolumn{2}{c}{Sphere} & \multicolumn{2}{c}{Rastrigin} & \multicolumn{2}{c}{Arm} & \multicolumn{2}{c}{Overcooked} \\
\cmidrule(lr){2-3} \cmidrule(lr){4-5} \cmidrule(lr){6-7} \cmidrule(lr){8-9}
 & MOQD-Score & Coverage & MOQD-Score & Coverage & MOQD-Score & Coverage & MOQD-Score & Coverage \\
\midrule
MO-CMA-MAE & \textbf{4,587,529.40} & \textbf{0.57} & \textbf{4,524,026.21} & \textbf{0.63} & \textbf{8,099,375.00} & \textbf{0.81} & \textbf{2,320,399.14} & \textbf{0.44} \\
MOME & 3,685,434.15 & 0.45 & 1,186,106.96 & 0.16 & 7,993,990.79 & 0.80 & 2,260,441.22 & \textbf{0.44} \\
NSGA-II & 1,021,801.49 & 0.13 & 230,325.90 & 0.03 & 6,790,374.55 & 0.68 & 1,935,600.67 & 0.39 \\
COMO-CMA-ES & 644,058.23 & 0.08 &   265,350.80 & 0.04 &   709,031.03 & 0.07 &    58,452.34 & 0.01 \\
SMS-EMOA & 883,143.82 & 0.11 & 818,285.22 & 0.11 & 6,452,321.40 & 0.65 & 1,538,610.23 & 0.33 \\
\bottomrule
\end{tabular}
}
\end{table*}

\begin{figure*}[!h]
    \centering
    \includegraphics[width=\linewidth]{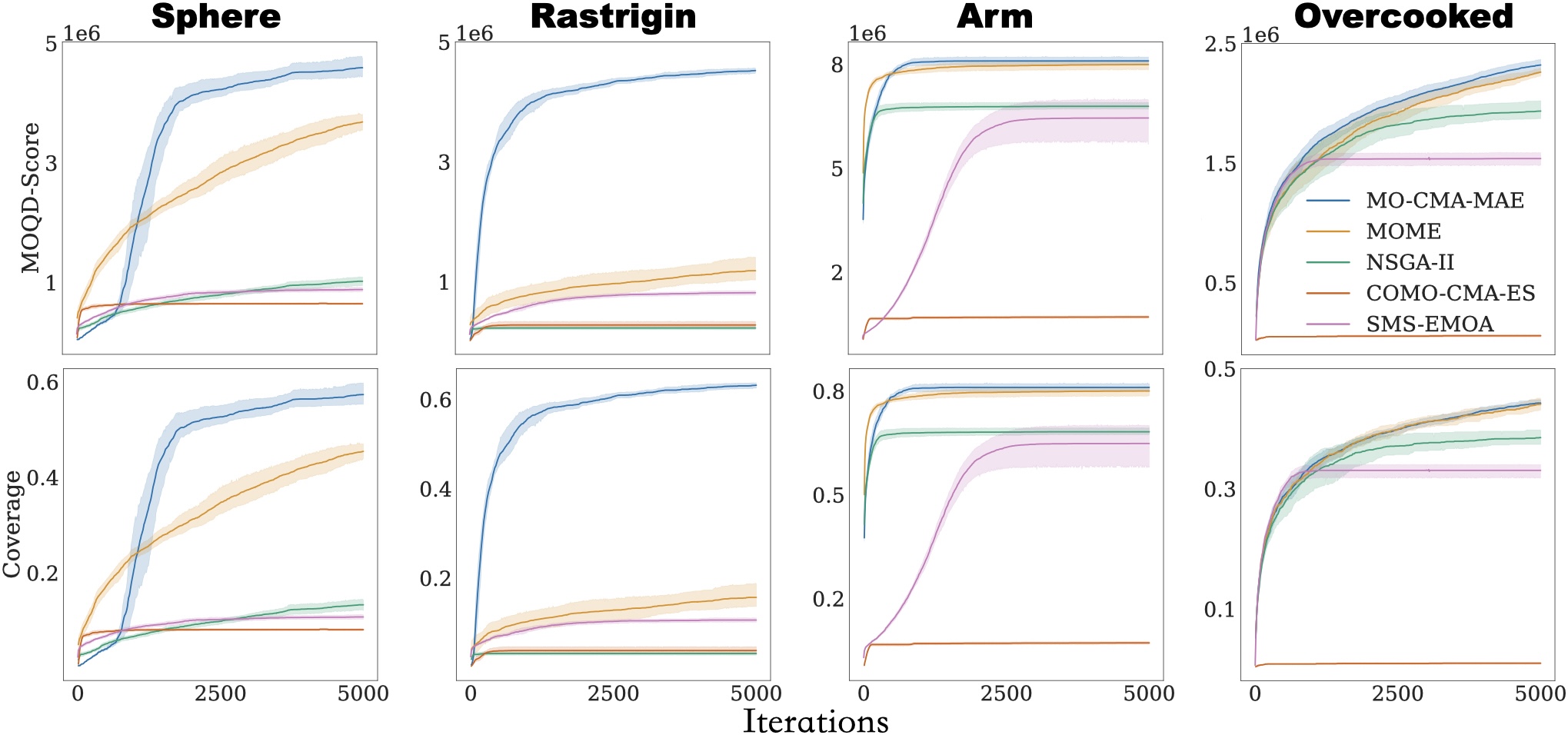}
    \caption{MOQD-scores and Coverages achieved by tested algorithms on each domain.}
    \label{fig:all}
\end{figure*}

\subsection{Main Results}
\label{sec:main_results}
We compare the MOQD-scores of MO-CMA-MAE and baseline algorithms in each domain with $5$ trials. The mean results are listed in Table~\ref{tab:results_new_domains}. To test the hypothesis that MO-CMA-MAE achieves significantly better MOQD-scores than baselines in each domain, we first conduct a one-way ANOVA test to examine whether there are significant variations across algorithms in each domain. Since all ANOVA test results were significant ($p < 0.001$), we performed pairwise comparisons with Tukey's HSD test for MOQD-scores grouped by algorithm in each domain. Results from the pairwise comparisons show that MO-CMA-MAE achieves significantly ($p < 0.05$) higher MOQD-scores than all baselines in the \textit{sphere} and \textit{rastrigin} domains. In the \textit{arm} and \textit{overcooked} domains, MO-CMA-MAE and MOME achieve similar MOQD-scores, and both outperform the other baselines. We plot the mean MOQD-scores and archive coverages with 95\% confidence intervals in Figure~\ref{fig:all}. For detailed statistical test results, please see Appendix~\ref{sec:statistical_test_results}.

\begin{figure*}[!h]
    \centering
    \includegraphics[width=\linewidth]{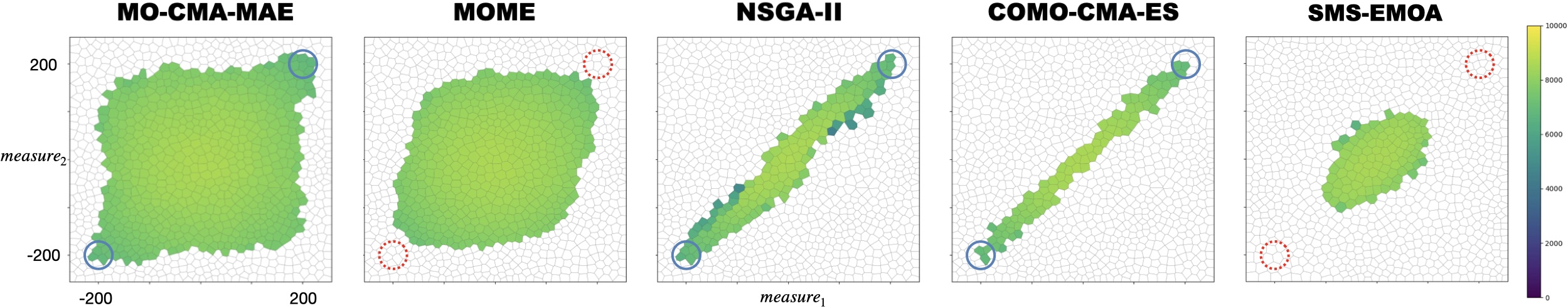}
    \caption{Heatmaps representing the passive MOQD archive populated by each algorithm after 5000 iterations of optimization on the \textit{sphere} domain. The color intensity within each cell corresponds to the hypervolume within it.}
    \label{fig:sphere_akv}
\end{figure*}

For deeper insights, we show in Figure~\ref{fig:sphere_akv} each baseline's passive MOQD archive heatmap for the \textit{sphere} domain. The color intensity of each cell corresponds to the hypervolume of its associated PF, with high intensity corresponding to large hypervolume, and blankness corresponding to empty cell. We focus on the \textit{sphere} domain because the
coupling between its objective and measure functions allows us to identify solutions with measure values along the line segment from $[-200, -200]$ to $[200, 200]$ as Pareto-optimal. We observe that, due to the lack of measure exploration incentives, NSGA-II, SMS-EMOA, COMO-CMA-ES focus exclusively on the archive cells along this line segment. In contrast, MOME explores more cells far away from this line segment, but fails to reach the endpoint cells at $[-200, -200]$ and $[200, 200]$ due to its reliance on non-adaptive search. Compared to these baselines, MO-CMA-MAE's utilization of adaptive CMA-ES search enables it to explore cells away from the line segment while also reaching both endpoint cells. We provide archive heatmaps for the remaining domains in Appendix~\ref{sec:other_akv}.

\subsection{Discussions on the Choice of $\alpha$}
\begin{figure}[h!]
    \centering
    \includegraphics[width=0.7\columnwidth]{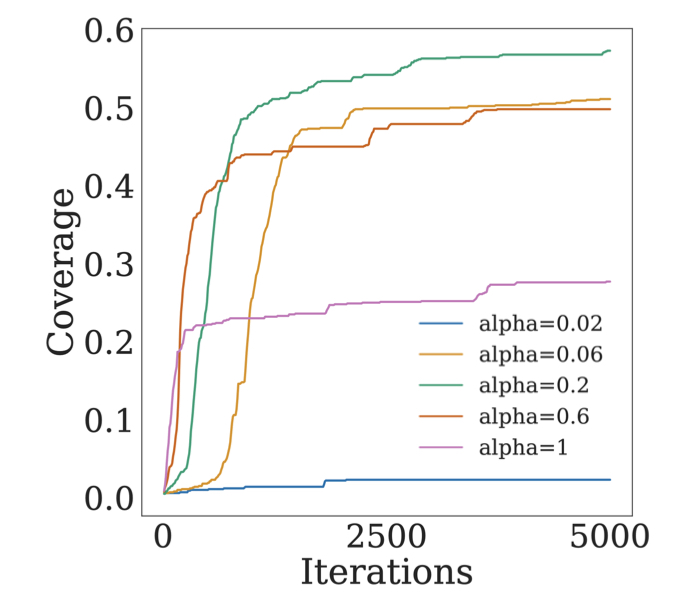}
    \caption{Sphere domain archive coverages at different $\alpha$.}
    \label{fig:alphas}
\end{figure}

In this section, we discuss the effects of the \textit{learning rate} hyperparameter $\alpha$ on MO-CMA-MAE measure space exploration. As explained in Section~\ref{sec:threshold_soft}, $\alpha$ controls the fraction of local gap between $T_e$ and $F_e$ that is closed with each insertion, and thereby the rate at which $T_e$ approximates $F_e$. With high $\alpha$ values, MO-CMA-MAE rapidly closes the gap between $T_e$ and $F_e$ within each cell $e$, quickly depleting the HVI boost for newly discovered archive cells. Conversely, very low $\alpha$ values keep $T_e$ far behind $F_e$, allowing MO-CMA-MAE to continue receiving positive HVI for cells that have already been well-explored. Both extremes cause poor measure-space exploration, but aside from extreme $\alpha$ values close to $0$ or $1$, we find MO-CMA-MAE to be reasonably robust to the choice of $\alpha$. We support this claim by referring to the archive coverages achieved with $\alpha$ = $\{0.02,$ $0.06,$ $0.2,$ $0.6,$ $1\}$, as shown in Figure~\ref{fig:alphas}. We note that setting $\alpha = 1$ disables MO-CMA-MAE's threshold accepting mechanism, so the poor performance corresponding to $\alpha = 1$ also demonstrates the importance of our threshold mechanism.

\begin{figure}[H]
    \centering
    \includegraphics[width=\linewidth]{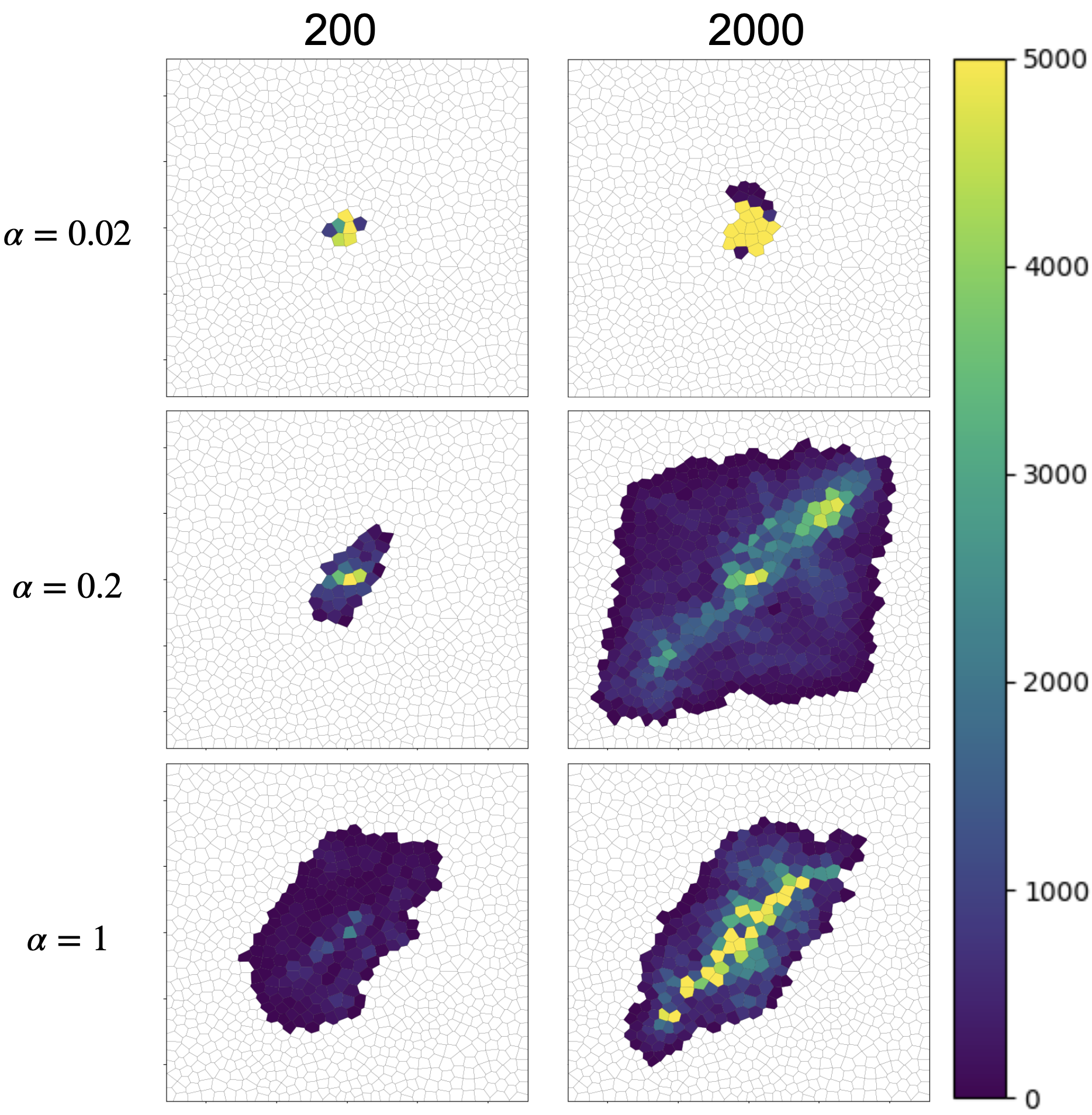}
    \caption{Frequency of CMA-ES samples from each cell at iterations $200$ and $2000$ when running on the \textit{sphere} domain.}
    \label{fig:numvisits}
\end{figure}

To further illustrate why $\alpha = 0.02$ and $\alpha = 1$ cause poor exploration, we show the frequency of CMA-ES samples from each cell in Figure~\ref{fig:numvisits}. When $\alpha = 0.02$, we observe that CMA-ES samples exclusively near the cell in which it is initialized. This is because even though these starting cells have received thousands of samples, $T_e$ remains far behind $F_e$ due to low $\alpha$ and continues boosting HVI rewards. As a result, CMA-ES continues receiving positive HVI for sampling within these starting cells, and has little incentive to shift its exploration to new cells. Conversely, when $\alpha = 1$, there is no HVI boost for sampling solutions within newly-discovered cells, so the HVI rewards within every discovered cell quickly depletes, forcing CMA-ES to constantly shift its sampling to new cells. Although this leads to fast exploration at the start, the depletion of HVI also makes it difficult for CMA-ES to receive any reward for further exploration, causing archive coverage to stagnate after the initial expansion.

\section{Conclusion and Future Work}
In this work, we propose MO-CMA-MAE, a novel MOQD algorithm that uses CMA-ES to learn to generate solutions that either explore new cells in measure space or improve incumbent PFs with already-discovered cells. We use a ranking defined by hypervolume improvement to guide CMA-ES updates. To encourage CMA-ES to explore the measure space, we compute hypervolume improvements relative to threshold fronts with each cell, which lag behind but gradually converge to their respective real PFs. Our experiments demonstrate that MO-CMA-MAE generally achieves better MOQD-score compared to other baselines.

Currently, hypervolume computation is the main bottleneck of MO-CMA-MAE. It is well-known that the computation time of hypervolume increases exponentially with the number of objectives~\citep{shang2020survey}, and MO-CMA-MAE frequently computes hypervolume improvement during bisection search and solution insertion. This limitation restricts its scalability to problems with only a few objectives. A potential solution to this bottleneck is to consider hypervolume approximation methods such as Monte Carlo approximation~\citep{bader2011hype} and surrogate models~\citep{shang2022hv} instead of computing hypervolume exactly. 

% However, with hypervolume approximation, it may be difficult to keep discount factor search error in the negative direction, which, as discussed in Section~\ref{sec:algorithm}, may cause the threshold front to skip over objective-space regions that haven't been covered. 

\newpage

\begin{acks}
This work has been partially supported by the Agilent Early Career Professor Award, the NSF CAREER \#2145077, NSF NRI \#2024949 and DARPA EMHAT HR00112490409. We thank Matthew C. Fontaine for helping improve our threshold accepting mechanism with his insights. We would also like to thank the anonymous reviewers for their feedback on a preliminary version of this paper.
\end{acks}

\bibliographystyle{plainnat}
\bibliography{Bibliography}

\begin{thebibliography}{50}
\providecommand{\natexlab}[1]{#1}
\providecommand{\url}[1]{\texttt{#1}}
\expandafter\ifx\csname urlstyle\endcsname\relax
  \providecommand{\doi}[1]{doi: #1}\else
  \providecommand{\doi}{doi: \begingroup \urlstyle{rm}\Url}\fi

\bibitem[Bader and Zitzler(2011)]{bader2011hype}
Johannes Bader and Eckart Zitzler.
\newblock Hype: An algorithm for fast hypervolume-based many-objective optimization.
\newblock \emph{Evolutionary computation}, 19\penalty0 (1):\penalty0 45--76, 2011.

\bibitem[Batra et~al.(2023)Batra, Tjanaka, Fontaine, Petrenko, Nikolaidis, and Sukhatme]{batra2023proximal}
Sumeet Batra, Bryon Tjanaka, Matthew~C Fontaine, Aleksei Petrenko, Stefanos Nikolaidis, and Gaurav Sukhatme.
\newblock Proximal policy gradient arborescence for quality diversity reinforcement learning.
\newblock \emph{arXiv preprint arXiv:2305.13795}, 2023.

\bibitem[Berghammer et~al.(2010)Berghammer, Friedrich, and Neumann]{berghammer2010set}
Rudolf Berghammer, Tobias Friedrich, and Frank Neumann.
\newblock Set-based multi-objective optimization, indicators, and deteriorative cycles.
\newblock In \emph{Proceedings of the 12th annual conference on Genetic and evolutionary computation}, pages 495--502, 2010.

\bibitem[Beume et~al.(2007)Beume, Naujoks, and Emmerich]{beume2007sms}
Nicola Beume, Boris Naujoks, and Michael Emmerich.
\newblock Sms-emoa: Multiobjective selection based on dominated hypervolume.
\newblock \emph{European Journal of Operational Research}, 181\penalty0 (3):\penalty0 1653--1669, 2007.

\bibitem[Bhatt et~al.(2022)Bhatt, Tjanaka, Fontaine, and Nikolaidis]{bhatt2022deep}
Varun Bhatt, Bryon Tjanaka, Matthew Fontaine, and Stefanos Nikolaidis.
\newblock Deep surrogate assisted generation of environments.
\newblock \emph{Advances in Neural Information Processing Systems}, 35:\penalty0 37762--37777, 2022.

\bibitem[{Blank} and {Deb}(2020)]{pymoo}
J.~{Blank} and K.~{Deb}.
\newblock pymoo: Multi-objective optimization in python.
\newblock \emph{IEEE Access}, 8:\penalty0 89497--89509, 2020.

\bibitem[Bringmann et~al.(2014)Bringmann, Friedrich, and Klitzke]{bringmann2014two}
Karl Bringmann, Tobias Friedrich, and Patrick Klitzke.
\newblock Two-dimensional subset selection for hypervolume and epsilon-indicator.
\newblock In \emph{Proceedings of the 2014 Annual Conference on Genetic and Evolutionary Computation}, pages 589--596, 2014.

\bibitem[Carroll et~al.(2019)Carroll, Shah, Ho, Griffiths, Seshia, Abbeel, and Dragan]{carroll2019utility}
Micah Carroll, Rohin Shah, Mark~K Ho, Tom Griffiths, Sanjit Seshia, Pieter Abbeel, and Anca Dragan.
\newblock On the utility of learning about humans for human-ai coordination.
\newblock \emph{Advances in neural information processing systems}, 32, 2019.

\bibitem[Chalumeau et~al.(2023)Chalumeau, Lim, Boige, Allard, Grillotti, Flageat, Macé, Flajolet, Pierrot, and Cully]{chalumeau2023qdax}
Felix Chalumeau, Bryan Lim, Raphael Boige, Maxime Allard, Luca Grillotti, Manon Flageat, Valentin Macé, Arthur Flajolet, Thomas Pierrot, and Antoine Cully.
\newblock Qdax: A library for quality-diversity and population-based algorithms with hardware acceleration, 2023.

\bibitem[Cully and Demiris(2017)]{cully2017quality}
Antoine Cully and Yiannis Demiris.
\newblock Quality and diversity optimization: A unifying modular framework.
\newblock \emph{IEEE Transactions on Evolutionary Computation}, 22\penalty0 (2):\penalty0 245--259, 2017.

\bibitem[Cully et~al.(2015)Cully, Clune, Tarapore, and Mouret]{cully2015robots}
Antoine Cully, Jeff Clune, Danesh Tarapore, and Jean-Baptiste Mouret.
\newblock Robots that can adapt like animals.
\newblock \emph{Nature}, 521\penalty0 (7553):\penalty0 503--507, 2015.

\bibitem[Deb et~al.(2002)Deb, Pratap, Agarwal, and Meyarivan]{deb2002fast}
Kalyanmoy Deb, Amrit Pratap, Sameer Agarwal, and TAMT Meyarivan.
\newblock A fast and elitist multiobjective genetic algorithm: Nsga-ii.
\newblock \emph{IEEE transactions on evolutionary computation}, 6\penalty0 (2):\penalty0 182--197, 2002.

\bibitem[D{\'\i}az-Manr{\'\i}quez et~al.(2013)D{\'\i}az-Manr{\'\i}quez, Toscano-Pulido, Coello, and Landa-Becerra]{diaz2013ranking}
Alan D{\'\i}az-Manr{\'\i}quez, Gregorio Toscano-Pulido, Carlos A~Coello Coello, and Ricardo Landa-Becerra.
\newblock A ranking method based on the r2 indicator for many-objective optimization.
\newblock In \emph{2013 IEEE Congress on Evolutionary Computation}, pages 1523--1530. IEEE, 2013.

\bibitem[Ding et~al.(2023)Ding, Zhang, Clune, Spector, and Lehman]{ding2023quality}
Li~Ding, Jenny Zhang, Jeff Clune, Lee Spector, and Joel Lehman.
\newblock Quality diversity through human feedback.
\newblock \emph{arXiv preprint arXiv:2310.12103}, 2023.

\bibitem[Fontaine and Nikolaidis(2023)]{fontaine2023covariance}
Matthew Fontaine and Stefanos Nikolaidis.
\newblock Covariance matrix adaptation map-annealing.
\newblock In \emph{Proceedings of the Genetic and Evolutionary Computation Conference}, pages 456--465, 2023.

\bibitem[Fontaine and Nikolaidis(2022)]{fontaine2022evaluating}
Matthew~C Fontaine and Stefanos Nikolaidis.
\newblock Evaluating human--robot interaction algorithms in shared autonomy via quality diversity scenario generation.
\newblock \emph{ACM Transactions on Human-Robot Interaction (THRI)}, 11\penalty0 (3):\penalty0 1--30, 2022.

\bibitem[Fontaine et~al.(2020)Fontaine, Togelius, Nikolaidis, and Hoover]{fontaine2020covariance}
Matthew~C Fontaine, Julian Togelius, Stefanos Nikolaidis, and Amy~K Hoover.
\newblock Covariance matrix adaptation for the rapid illumination of behavior space.
\newblock In \emph{Proceedings of the 2020 genetic and evolutionary computation conference}, pages 94--102, 2020.

\bibitem[Fontaine et~al.(2021)Fontaine, Hsu, Zhang, Tjanaka, and Nikolaidis]{fontaine2021importance}
Matthew~C Fontaine, Ya-Chuan Hsu, Yulun Zhang, Bryon Tjanaka, and Stefanos Nikolaidis.
\newblock On the importance of environments in human-robot coordination.
\newblock \emph{arXiv preprint arXiv:2106.10853}, 2021.

\bibitem[Freeman et~al.(2021)Freeman, Frey, Raichuk, Girgin, Mordatch, and Bachem]{freeman2021brax}
C~Daniel Freeman, Erik Frey, Anton Raichuk, Sertan Girgin, Igor Mordatch, and Olivier Bachem.
\newblock Brax--a differentiable physics engine for large scale rigid body simulation.
\newblock \emph{arXiv preprint arXiv:2106.13281}, 2021.

\bibitem[Gallotta et~al.(2023)Gallotta, Arulkumaran, and Soros]{gallotta2023preference}
Roberto Gallotta, Kai Arulkumaran, and Lisa~B Soros.
\newblock Preference-learning emitters for mixed-initiative quality-diversity algorithms.
\newblock \emph{IEEE Transactions on Games}, 2023.

\bibitem[GhostTownGames(2016)]{overcooked}
GhostTownGames.
\newblock Overcooked, 2016.
\newblock URL \url{https://store.steampowered.com/app/ 448510/Overcooked/}.

\bibitem[Goodfellow et~al.(2014)Goodfellow, Pouget-Abadie, Mirza, Xu, Warde-Farley, Ozair, Courville, and Bengio]{goodfellow2014generative}
Ian Goodfellow, Jean Pouget-Abadie, Mehdi Mirza, Bing Xu, David Warde-Farley, Sherjil Ozair, Aaron Courville, and Yoshua Bengio.
\newblock Generative adversarial nets.
\newblock \emph{Advances in neural information processing systems}, 27, 2014.

\bibitem[Grillotti and Cully(2022)]{grillotti2022unsupervised}
Luca Grillotti and Antoine Cully.
\newblock Unsupervised behavior discovery with quality-diversity optimization.
\newblock \emph{IEEE Transactions on Evolutionary Computation}, 26\penalty0 (6):\penalty0 1539--1552, 2022.

\bibitem[Hansen(2016)]{hansen:cma16}
Nikolaus Hansen.
\newblock The cma evolution strategy: A tutorial.
\newblock \emph{arXiv preprint arXiv:1604.00772}, 2016.

\bibitem[Hansen(2020)]{hansenmoarchiving}
Nikolaus Hansen.
\newblock Moarchiving.
\newblock \url{https://github.com/CMA-ES/moarchiving/tree/master}, 2020.

\bibitem[Hansen et~al.(2021)Hansen, Auger, Ros, Mersmann, Tu{\v{s}}ar, and Brockhoff]{hansen2021coco}
Nikolaus Hansen, Anne Auger, Raymond Ros, Olaf Mersmann, Tea Tu{\v{s}}ar, and Dimo Brockhoff.
\newblock Coco: A platform for comparing continuous optimizers in a black-box setting.
\newblock \emph{Optimization Methods and Software}, 36\penalty0 (1):\penalty0 114--144, 2021.

\bibitem[Igel et~al.(2007)Igel, Hansen, and Roth]{igel2007covariance}
Christian Igel, Nikolaus Hansen, and Stefan Roth.
\newblock Covariance matrix adaptation for multi-objective optimization.
\newblock \emph{Evolutionary computation}, 15\penalty0 (1):\penalty0 1--28, 2007.

\bibitem[Ishibuchi et~al.(2015)Ishibuchi, Masuda, Tanigaki, and Nojima]{ishibuchi2015modified}
Hisao Ishibuchi, Hiroyuki Masuda, Yuki Tanigaki, and Yusuke Nojima.
\newblock Modified distance calculation in generational distance and inverted generational distance.
\newblock In \emph{Evolutionary Multi-Criterion Optimization: 8th International Conference, EMO 2015, Guimar{\~a}es, Portugal, March 29--April 1, 2015. Proceedings, Part II 8}, pages 110--125. Springer, 2015.

\bibitem[Janmohamed et~al.(2023)Janmohamed, Pierrot, and Cully]{janmohamed2023improving}
Hannah Janmohamed, Thomas Pierrot, and Antoine Cully.
\newblock Improving the data efficiency of multi-objective quality-diversity through gradient assistance and crowding exploration.
\newblock In \emph{Proceedings of the Genetic and Evolutionary Computation Conference}, pages 165--173, 2023.

\bibitem[Jiang et~al.(2014)Jiang, Zhang, Ong, Zhang, and Tan]{jiang2014simple}
Siwei Jiang, Jie Zhang, Yew-Soon Ong, Allan~N Zhang, and Puay~Siew Tan.
\newblock A simple and fast hypervolume indicator-based multiobjective evolutionary algorithm.
\newblock \emph{IEEE Transactions on Cybernetics}, 45\penalty0 (10):\penalty0 2202--2213, 2014.

\bibitem[Littman et~al.(1995)Littman, Cassandra, and Kaelbling]{littman1995learning}
Michael~L Littman, Anthony~R Cassandra, and Leslie~Pack Kaelbling.
\newblock Learning policies for partially observable environments: Scaling up.
\newblock In \emph{Machine Learning Proceedings 1995}, pages 362--370. Elsevier, 1995.

\bibitem[Menchaca-Mendez and Coello~Coello(2015)]{menchaca2015gd}
Adriana Menchaca-Mendez and Carlos~A Coello~Coello.
\newblock Gd-moea: A new multi-objective evolutionary algorithm based on the generational distance indicator.
\newblock In \emph{International conference on evolutionary multi-criterion optimization}, pages 156--170. Springer, 2015.

\bibitem[Mouret and Clune(2015)]{mouret2015illuminating}
Jean-Baptiste Mouret and Jeff Clune.
\newblock Illuminating search spaces by mapping elites.
\newblock \emph{arXiv preprint arXiv:1504.04909}, 2015.

\bibitem[Nilsson and Cully(2021)]{nilsson2021policy}
Olle Nilsson and Antoine Cully.
\newblock Policy gradient assisted map-elites.
\newblock In \emph{Proceedings of the Genetic and Evolutionary Computation Conference}, pages 866--875, 2021.

\bibitem[Pierrot et~al.(2022)Pierrot, Richard, Beguir, and Cully]{pierrot2022multi}
Thomas Pierrot, Guillaume Richard, Karim Beguir, and Antoine Cully.
\newblock Multi-objective quality diversity optimization.
\newblock In \emph{Proceedings of the Genetic and Evolutionary Computation Conference}, pages 139--147, 2022.

\bibitem[Purshouse and Fleming(2007)]{purshouse2007evolutionary}
Robin~C Purshouse and Peter~J Fleming.
\newblock On the evolutionary optimization of many conflicting objectives.
\newblock \emph{IEEE transactions on evolutionary computation}, 11\penalty0 (6):\penalty0 770--784, 2007.

\bibitem[Shang et~al.(2020)Shang, Ishibuchi, He, and Pang]{shang2020survey}
Ke~Shang, Hisao Ishibuchi, Linjun He, and Lie~Meng Pang.
\newblock A survey on the hypervolume indicator in evolutionary multiobjective optimization.
\newblock \emph{IEEE Transactions on Evolutionary Computation}, 25\penalty0 (1):\penalty0 1--20, 2020.

\bibitem[Shang et~al.(2022)Shang, Chen, Liao, and Ishibuchi]{shang2022hv}
Ke~Shang, Weiyu Chen, Weiduo Liao, and Hisao Ishibuchi.
\newblock Hv-net: Hypervolume approximation based on deepsets.
\newblock \emph{IEEE Transactions on Evolutionary Computation}, 27\penalty0 (4):\penalty0 1154--1160, 2022.

\bibitem[Tian et~al.(2017)Tian, Cheng, Zhang, Cheng, and Jin]{tian2017indicator}
Ye~Tian, Ran Cheng, Xingyi Zhang, Fan Cheng, and Yaochu Jin.
\newblock An indicator-based multiobjective evolutionary algorithm with reference point adaptation for better versatility.
\newblock \emph{IEEE Transactions on Evolutionary Computation}, 22\penalty0 (4):\penalty0 609--622, 2017.

\bibitem[Tjanaka et~al.(2022)Tjanaka, Fontaine, Togelius, and Nikolaidis]{tjanaka2022approximating}
Bryon Tjanaka, Matthew~C Fontaine, Julian Togelius, and Stefanos Nikolaidis.
\newblock Approximating gradients for differentiable quality diversity in reinforcement learning.
\newblock In \emph{Proceedings of the Genetic and Evolutionary Computation Conference}, pages 1102--1111, 2022.

\bibitem[Tjanaka et~al.(2023)Tjanaka, Fontaine, Lee, Zhang, Balam, Dennler, Garlanka, Klapsis, and Nikolaidis]{pyribs}
Bryon Tjanaka, Matthew~C Fontaine, David~H Lee, Yulun Zhang, Nivedit~Reddy Balam, Nathaniel Dennler, Sujay~S Garlanka, Nikitas~Dimitri Klapsis, and Stefanos Nikolaidis.
\newblock pyribs: A bare-bones python library for quality diversity optimization.
\newblock In \emph{Proceedings of the Genetic and Evolutionary Computation Conference}, GECCO '23, page 220–229, New York, NY, USA, 2023. Association for Computing Machinery.
\newblock ISBN 9798400701191.
\newblock \doi{10.1145/3583131.3590374}.
\newblock URL \url{https://doi.org/10.1145/3583131.3590374}.

\bibitem[Tour{\'e} et~al.(2019)Tour{\'e}, Hansen, Auger, and Brockhoff]{toure2019uncrowded}
Cheikh Tour{\'e}, Nikolaus Hansen, Anne Auger, and Dimo Brockhoff.
\newblock Uncrowded hypervolume improvement: Como-cma-es and the sofomore framework.
\newblock In \emph{Proceedings of the Genetic and Evolutionary Computation Conference}, pages 638--646, 2019.

\bibitem[Trautmann et~al.(2013)Trautmann, Wagner, and Brockhoff]{trautmann2013r2}
Heike Trautmann, Tobias Wagner, and Dimo Brockhoff.
\newblock R2-emoa: Focused multiobjective search using r2-indicator-based selection.
\newblock In \emph{Learning and Intelligent Optimization: 7th International Conference, LION 7, Catania, Italy, January 7-11, 2013, Revised Selected Papers 7}, pages 70--74. Springer, 2013.

\bibitem[Vassiliades and Mouret(2018)]{vassiliades2018discovering}
Vassiiis Vassiliades and Jean-Baptiste Mouret.
\newblock Discovering the elite hypervolume by leveraging interspecies correlation.
\newblock In \emph{Proceedings of the Genetic and Evolutionary Computation Conference}, pages 149--156, 2018.

\bibitem[Vassiliades et~al.(2016)Vassiliades, Chatzilygeroudis, and Mouret]{vassiliades2016scaling}
Vassilis Vassiliades, Konstantinos Chatzilygeroudis, and Jean-Baptiste Mouret.
\newblock Scaling up map-elites using centroidal voronoi tessellations.
\newblock \emph{arXiv preprint arXiv:1610.05729}, 2016.

\bibitem[Vo{\ss} et~al.(2009)Vo{\ss}, Hansen, and Igel]{voss2009recombination}
Thomas Vo{\ss}, Nikolaus Hansen, and Christian Igel.
\newblock Recombination for learning strategy parameters in the mo-cma-es.
\newblock In \emph{International Conference on Evolutionary Multi-Criterion Optimization}, pages 155--168. Springer, 2009.

\bibitem[Vo{\ss} et~al.(2010)Vo{\ss}, Hansen, and Igel]{voss2010improved}
Thomas Vo{\ss}, Nikolaus Hansen, and Christian Igel.
\newblock Improved step size adaptation for the mo-cma-es.
\newblock In \emph{Proceedings of the 12th annual conference on Genetic and evolutionary computation}, pages 487--494, 2010.

\bibitem[Zhao et~al.(2024)Zhao, Tjanaka, Fontaine, and Nikolaidis]{zhaocovariance}
Shihan Zhao, Bryon Tjanaka, Matthew~C Fontaine, and Stefanos Nikolaidis.
\newblock Covariance matrix adaptation map-annealing: Theory and experiments.
\newblock \emph{ACM Transactions on Evolutionary Learning}, 2024.

\bibitem[Zitzler and K{\"u}nzli(2004)]{zitzler2004indicator}
Eckart Zitzler and Simon K{\"u}nzli.
\newblock Indicator-based selection in multiobjective search.
\newblock In \emph{International conference on parallel problem solving from nature}, pages 832--842. Springer, 2004.

\bibitem[Zitzler and Thiele(1999)]{zitzler1999multiobjective}
Eckart Zitzler and Lothar Thiele.
\newblock Multiobjective evolutionary algorithms: a comparative case study and the strength pareto approach.
\newblock \emph{IEEE transactions on Evolutionary Computation}, 3\penalty0 (4):\penalty0 257--271, 1999.

\end{thebibliography}

\clearpage
\appendix
\section{Hyperparameyer Selection}
\label{sec:hyperparams}
\noindent\textbf{Sphere, Rastrigin, Arm}
\begin{itemize}
    \item MO-CMA-MAE: $\sigma_0 = 0.5$, \textit{improvement} ranker~\citep{fontaine2020covariance}, $\mu$ selection, \textit{cycle} restart rule
    \item MOME: $\sigma_{iso} = 0.05$, $\sigma_{line} = 0.5$
    \item NSGA-II: $\sigma_{iso} = 0.05$, $\sigma_{line} = 0.5$
    \item COMO-CMA-ES: $\sigma_0 = 0.5$, \textit{improvement} ranker, $\mu$ selection, \textit{basic} restart rule
\end{itemize}

\noindent\textbf{Overcooked}
\begin{itemize}
    \item MO-CMA-MAE: $\sigma_0 = 0.5$, \textit{improvement} ranker, \textit{filter} selection, \textit{cycle} restart rule
    \item MOME: $\sigma_{iso} = 0.05$, $\sigma_{line} = 0.5$
    \item NSGA-II: $\sigma_{iso} = 0.05$, $\sigma_{line} = 0.5$
    \item COMO-CMA-ES: $\sigma_0 = 0.5$, \textit{improvement} ranker, \textit{filter} selection, \textit{basic} restart rule
\end{itemize}

\textit{Basic} restart rule triggers when one of the \textbf{ConditionCov}, \textbf{Stagnation}, or \textbf{TolXUp} CMA-ES convergence conditions~\citep{hansen:cma16} is met. \textit{Cycle} restart was described in Section~\ref{sec:downsize}. We switched from $\mu$ to \textit{filter} selection~\citep{pyribs} in the overcooked domain because sometimes over half of the candidate solutions may fail to evaluate due to high variance.

\section{Bisection Search Implementation}
\label{sec:bisec}
Given the objective values of a candidate solution ${\bf f}(\bm{x})$, the current threshold front $T_e$, the MO-CMA-MAE \textit{learning rate} $\alpha$, and the search error tolerance $\epsilon$, the $\mbox{bisect}({\bf f}(\bm{x}), T_e, \alpha, \epsilon)$ function returns a discount factor $d$ satisfying $\mbox{HVI}_r(d{\bf f}(\bm{x}, T_e)) \approx \alpha\mbox{HVI}_r({\bf f}(\bm{x}, T_e))$. We highlight that we restrict the error to be in the negative direction, i.e $\mbox{HVI}_r(d{\bf f}(\bm{x}), T_e) \leq \alpha \mbox{HVI}_r({\bf f}(\bm{x}), T_e)$, due to the observation that the outward expansion of $T_e$ is non-reversible. Specifically, inserting a large $d{\bf f}(\bm{x})$ will prevent $T_e$ from accepting any future solutions dominated by it, effectively causing $T_e$ ``skip over'' some regions of the objective space that it has not actually covered. We show the pseudocodes of our bisection search implementation in Algorithm~\ref{alg:bisec}.

\begin{algorithm}[h]
\SetAlgoLined
\caption{Bisection Search}
\label{alg:bisec}
\SetKwInOut{Input}{input}
\SetKwInOut{Result}{result}
\SetKwProg{bisect}{bisect}{}{}
\SetKwRepeat{Do}{do}{while}
\DontPrintSemicolon
\bisect{$({\bf f}(\bm{x}), T_e, \alpha, \epsilon)$}
{

\Input{The objective values of the candidate solution to be inserted ${\bf f}(\bm{x})$, the current threshold front $T_e$, the learning rate $\alpha$, the search error tolerance $\epsilon$.}
\Result{The discount factor to be applied to ${\bf f}(\bm{x})$ before inserting to $T_e$.}

\BlankLine

$target\_hvi = \alpha \mbox{HVI}_r({\bf f}(\bm{x}), T_e)$\;
$low = 0$\;
$high = 1$\;

\BlankLine

\Do{$mid\_hvi > target$ \textbf{or} $target - mid\_hvi > \epsilon$}{
    $d = (low + high) / 2$\;
    $mid\_hvi = \mbox{HVI}_r(d{\bf f}(\bm{x}), T_e)$\;

    \If{$mid\_hvi > target$}{
        \tcc{Search down if current hvi is larger than target}
        $hi \gets d_i$\;
    }
    \Else {
        \tcc{Search up if current hvi is smaller than target by more than $\epsilon$}
        $lo \gets d_i$\;
    } 
}

\BlankLine

\Return $d$\;
}

\end{algorithm}

\section{Emitter COMO-CMA-ES}
\label{sec:comocmaes}
For the COMO-CMA-ES baseline, we adapt the original algorithm to incorporate an emitter architecture to make it more compatible with the other baselines. The original COMO-CMA-ES maintains a small number of CMA-ES instances as its population, and uses the mean state maintained within every instance to construct the PF. Given the large population size we use across all baselines, maintaining thousands of CMA-ES instances and using their means for the PF is impractical. Instead, we maintain a small number of CMA-ES instances, and use them only as emitters for sampling solutions. The non-dominated solutions from these samples are retained as both the PF and the population of COMO-CMA-ES. We note that, compared to the modified COMO-CMA-ES, MO-CMA-MAE adds explicit measure space quantification, and does not augment the HVI indicator with an additional condition for dominated solutions. These differences provide us with an opportunity to assess the effectiveness of MO-CMA-MAE's diversifying and threshold mechanisms compared to another MOO algorithm that is also driven by CMA-ES and hypervolume-based indicator. We show the pseudocodes for our modified Emitter-COMO-CMA-ES in Algorithm~\ref{alg:emitter_como_cma_es}.

\begin{algorithm}[h!]
\SetAlgoLined
\caption{Emitter-COMO-CMA-ES}
\label{alg:emitter_como_cma_es}
\SetKwInOut{Input}{input}
\SetKwInOut{Result}{result}
\SetKwProg{ECOMOCMAES}{Emitter-COMO-CMA-ES}{}{}
\DontPrintSemicolon
\ECOMOCMAES{$(N, \psi, \lambda, \bm{x}_0, \sigma_0)$}
{

\Input{A desired number of iterations $N$, the number of emitters $\psi$, a branching population size $\lambda$, an initial solution $\bm{x}_0$, and an initial step size $\sigma_0$.}
\Result{A global Pareto Set $P$ consisted of mutually non-dominated solutions.}

\BlankLine
Initialize $P \gets \emptyset\; F \gets \emptyset$.

Initialize $\psi$ CMA-ES emitters, each with initial mean $\bm{\bar{x}} \leftarrow \bm{x}_0$, initial covariance matrix $\mathcal{C} \leftarrow \sigma_0\mathbf{I}$, and other internal parameters $\bm{p}$.

\For{$iter\leftarrow 1$ \KwTo $N$}{
    \For{$i\leftarrow 1$ \KwTo $\lambda$}{
        $\bm{x}_i \sim \mathcal{N}(\bm{\bar{x}},\mathcal{C})$\; 
        
        ${\bf f}(\bm{x}_i) \gets \mbox{evaluate}(\bm{x}_i)$\;
        
        \BlankLine
        
        \If{$\bm{x}_i \not\prec P$}{
            \tcc{If $\bm{x}_i$ is non-dominated, compute its hypervolume improvement relative to the incumbent PF for later CMA-ES ranking}
            $\Phi_i \gets \mbox{HVI}_r({\bf f}(\bm{x}_i), F)$\; 
            \tcc{Add $\bm{x}_i$ to the incumbent PS and PF}
            $P \gets \{\bm{x}_i\} \cup \{\bm{x} \in P \mid \bm{x}_i \not\succ \bm{x}\}$\; 
            $F \gets \{{\bf f}(x) \; \forall x \in P \}$\; 
        }
        \Else {
            \tcc{If $\bm{x}_i$ is dominated, compute its negated distance to the incumbent PF for later CMA-ES ranking}
            $\Phi_i \gets -\mbox{dist}_r({\bf f}(\bm{x}_i), F)$\; 
        }

    \BlankLine
    \tcc{Adapt CMA-ES according to $\Phi_i$}
    rank $\bm{x}_i$ by $\Phi_i$ \; 
    adapt CMA-ES parameters $\bm{\bar{x}},\mathcal{C},\bm{p}$\; \label{mocmamae:adaptation}
    \If{\mbox{CMA-ES converges}}{
        $\bm{x}_0 \gets \mbox{sample\_elite}(P)$\;
        Restart CMA-ES with $\mathcal{C} \leftarrow \sigma_0 I$ and $\bm{\bar{x}} \leftarrow \bm{x}_0$.
    }
}
}
}

\end{algorithm}

\section{Domains}
\label{sec:domains}

We describe the objective and measure functions of our \textit{sphere}, \textit{rastrigin}, \textit{arm}, and \textit{overcooked} domains in this section. 

\noindent\textbf{Sphere and Rastrigin.} We modify the sphere domain from prior work~\citep{fontaine2020covariance, hansen2021coco} to have two shifted extrema. The bi-objective $\textbf{f}_{sphere}$ function is defined as:

\begin{equation}
    \label{eq:sphere}
    \textbf{f}_{sphere}(\bm{x} )= 
    \begin{bmatrix}
        -\Sigma_{i=1}^n[(x_i - \lambda_1)^2] \\
        -\Sigma_{i=1}^n[(x_i - \lambda_2)^2]
    \end{bmatrix}
\end{equation}

We choose $n=100$, $\lambda_1 = 4$, and $\lambda_2 = -4$ for our experiments. Same as prior work, we divide the parameters into two halves, and calculate $m_1({\bm{x}}) = \Sigma_{i \in [0:\lfloor n/2 \rfloor]}clip(x_i)$ and $m_2(\bm{x}) = \Sigma_{i \in [\lceil n/2 \rceil:n]}clip(x_i)$ as our measure functions. The $clip$ function serves to restrict the contribution of each $x_i$ to $[-5.12, 5.12]^n$, and is defined as:

\begin{equation}
    \label{eq:clip}
    clip(x_i) =
    \begin{cases}
        x_i & \text{if $-5.12 \leq x_i \leq 5.12$} \\
        5.12 / x_i & \text{otherwise}
    \end{cases}
\end{equation}

Given our choice of $\lambda_1, \lambda_2$, any solution that does not fall on the line segment going from $x_{1,2,\ldots,n}=-4$ to $x_{1,2,\ldots,n}=4$ can reduce its distance to both extrema by projecting onto this line segment, and thus cannot be Pareto-optimal. In other words, Pareto-optimal solutions should satisfy $-4 \leq x_1=x_2=\ldots=x_n \leq 4$, which means $m_1({\bm{x}}) = m_2({\bm{x}})$. Because of this, the \textit{sphere} domain strongly encourages optimizing cells along the primary diagonal in measure space, where $m_1(\bm{x}) \approx m_2(\bm{x})$. On the other hand, cells that are far away from this line segment are less optimizable, and require a diversifying mechanism to explore.

Similarly, we modify the $n=100$ rastrigin domain to have two shifted extrema at $\lambda_1 = 4$, and $\lambda_2 = -4$, defined as:

\begin{equation}
    \label{eq:rastrigin}
    \textbf{f}_{rastrigin}(\bm{x}) = 
    \begin{bmatrix}
        \Sigma_{i=1}^n[10 \cos (2\pi (x_i - \lambda_1)) - (x_i - \lambda_1)^2] \\
        \Sigma_{i=1}^n[10 \cos (2\pi (x_i - \lambda_2)) - (x_i - \lambda_2)^2]
    \end{bmatrix}
\end{equation}

We again use $clip$ as our measure function. This creates several optima along the primary diagonal within measure space, and similar to the sphere domain, requires a diversifying mechanism to explore off the primary diagonal.

In order to ensure a worst-case reference point $r$ for hypervolume calculation, we map all objective values to the range $[0,100]$ using min-max normalization, and set $r = {\bf 0}$. The lowest possible value for each objective is $0$ in both sphere and rastrigin domains. To get upper bounds on the objectives, we observe that for the sphere domain, $\max_{x_i \in [-10.24,10.24]}\{\textbf{f}_{sphere}(\bm{x})\} = 3893.76$; for the rastrigin domain, we use grid search to get an approximate upper bound $\max_{x_i \in [-10.24,10.24]}\{\textbf{f}_{rastrigin}(\bm{x})\} \approx 20214.97$. Therefore, we map the sphere domain's objectives from $[0, 3893.76]$ to $[0, 100]$, and rastrigin's objectives from $[0, 20214.97]$ to $[0, 100]$. The parameter bound $x_i \in [-10.24,10.24]$ is chosen to be weaker than the $[-5.12, 5.12]^n$ bounds restricted by the $clip$ measure function.

\begin{figure*}[t]
    \centering
    \includegraphics[width=0.8\linewidth]{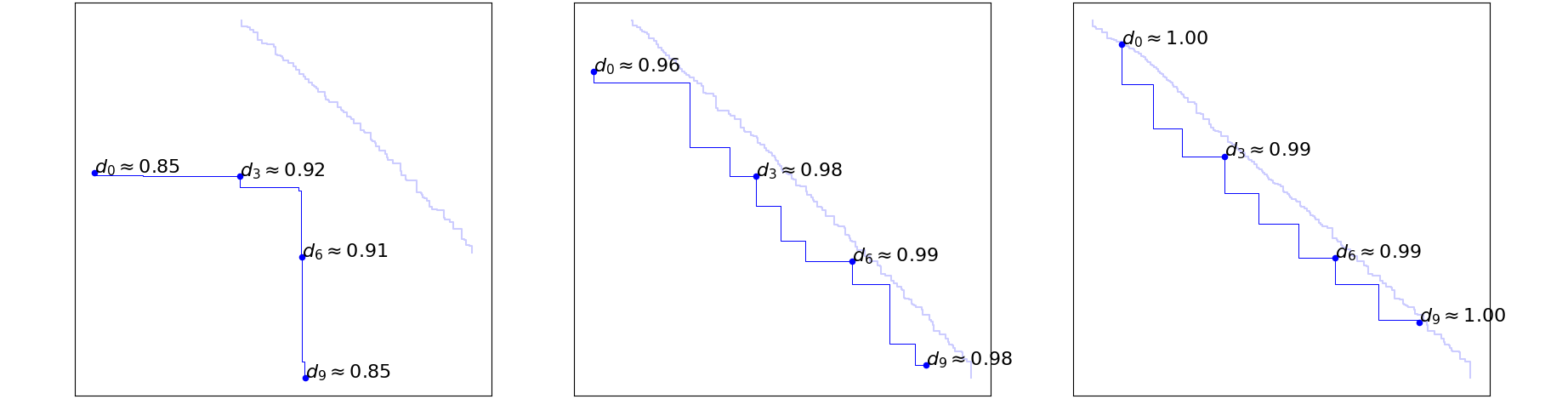}
    \caption{An example of how the threshold front approximates the real front in the static archive version of MO-CMA-MAE with size limit $10$. The three snapshots are taken from a specific archive cell at iterations $30$, $50$ and $1000$ when running on the sphere domain.}
    \label{fig:static_approx}
\end{figure*}

\begin{figure*}[t]
    \centering
    \includegraphics[width=0.8\linewidth]{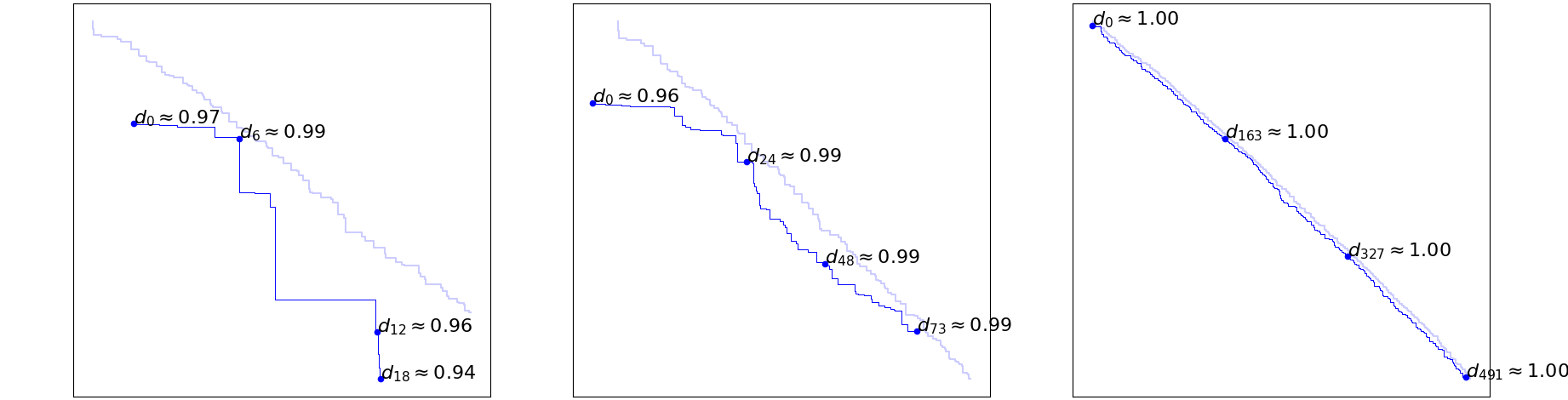}
    \caption{An example of how the threshold front approximates the real front in the dynamic archive version of MO-CMA-MAE. The three snapshots are taken from a specific archive cell at iterations $30$, $50$ and $1000$ when running on the sphere domain.}
    \label{fig:dynamic_approx}
\end{figure*}

\noindent\textbf{Arm Repertoire.} We modify the \textit{arm} domain from prior work~\citep{cully2017quality}. This domain simulates an $n$-DOF planar arm with revolute joints, and aims to find $n$ joint angles with the smallest variance that put the arm tip at a diverse set of locations. To obtain two non-aligned objectives, we define the bi-objective function to be the respective variances among the first and second halves of the joint angles:

\begin{equation}
    \label{eq:arm}
    \textbf{f}_{arm}(\bm{x}) = 
    \begin{bmatrix}
        -\mbox{Var}(x_{1}, \ldots, x_{\lfloor n/2 \rfloor}) \\
        -\mbox{Var}(x_{\lceil n/2 \rceil}, \ldots, x_{n})
    \end{bmatrix}
\end{equation}

We search over $n=100$ and $x_i \in [-\pi, \pi]$. We again use min-max normalization to map the objectives from $[-6.58, 0]$ to $[0, 100]$, where $-6.58$ corresponds to two times the expected variance when $n=100$ and $x_i \in [-\pi, \pi]$. In the rare event that a solution has larger variances (smaller negated variances), it is treated as a failed evaluation and discarded. The measures are the Cartesian coordinates of the arm tip computed with forward kinematics. The feasible measure space is a circle with radius = $nL$, where $L=1$ is the length of each arm link.

\noindent\textbf{Overcooked Level Generation.} Overcooked-AI~\citep{carroll2019utility} is an open source domain based on the popular video game \emph{Overcooked}~\citep{overcooked} for researching the coordination of agent behaviors. The domain features a tile-based map of a simulated kitchen, with key locations such as ingredients and the cooking pot dispersed across the map. In order to cook and serve dishes as quickly as possible, two cooperating agents need to effectively distribute tasks between themselves, and navigate to key locations while avoiding walls and each other. Prior work~\citep{fontaine2021importance} has used QD algorithms to generate maps inducing diverse player strategies for the overcooked-AI domain. In the aforementioned work, a generative adversarial network (GAN)~\citep{goodfellow2014generative} was trained to generate likely maps using human-authored examples, and QD algorithms were used to search over this GAN's latent space to generate diverse maps, which were finally repaired by mixed-integer linear programs (MIPs) to ensure the generated maps were playable.

In this work, we use a similar pipeline to generate diverse overcooked levels that are difficult and/or cheap to construct. We consider two objectives to be optimized: (1) \textit{difficulty}, which is the negated reward received by a pair of pre-defined agents, and (2) \textit{emptiness}, which is represented by the proportion of map tiles that are empty. We note that these two objectives are non-aligned, since difficult maps tend to have more walls. Parallel to the two non-aligned objectives, we explore over a measure space defined by two measure functions from the aforementioned work. These are (1) the difference between the numbers of ingredients and (2) the difference between the numbers of plates, picked up by the two agents. The measures represent how the agents distribute the workloads between themselves.

Two agents simulating a pair of cooperating players are used to evaluate the difficulty and workload distributions of a map. One is a rule-based agent that myopically selects the highest priority subtask based on the world state. The other is a QMDP~\citep{littman1995learning} planner that anticipates the first agent's plan and responds accordingly. The agents may choose random actions to unstuck if they run into each other, so the evaluation results are stochastic. Therefore, $4$ trials are run for each evaluation, and the median difficulty and mean measures are taken as the aggregate evaluation result. Emptiness is non-stochastic and is computed directly from the map. Even though averaging the measures may result in unrealistic measure values, such as a $0.5$ difference between the numbers of ingredients held. We interpret this as representing a stochastic margin when, for example, a map is equally likely to induce measure values $1$ and $0$. We also check the variances across evaluation results, and if the variance of either measure exceeds $0.5$, the evaluation for this map is failed, and is ignored by the tested algorithm.

All generated maps have $15 \times 10$ tiles, and evaluation of a map completes after either $100$ timesteps have passed, or $3$ dishes have been delivered to the counter. Game reward is computed as:

\begin{equation}
    \label{eq:reward}
    reward = (20 - \frac{t_1}{10}) b_1 + (20 - \frac{t_2}{10}) b_2 + (20 - \frac{t_3}{10}) b_3
\end{equation}

where $b_i$ is a binary indicating whether a dish has been delivered, and $t_i$ is the time taken to deliver a dish since the previous deliverance ($0$ in the case of first deliverance). The \textit{difficulty} and \textit{emptiness} objective scores are mapped to the range $[0, 100]$ using min-max normalization. Implementations for the agents and the MIP repair module are the same as in \citet{fontaine2021importance}.

\section{Additional Results}
\subsection{Threshold Front Convergence}
\label{sec:tpf_converge}
For visual intuition, we show two examples of how $T_e$ approximates $F_e$ when running MO-CMA-MAE with $\alpha = 0.1$ on the sphere domain. In Figure~\ref{fig:static_approx}, $T_e$ is constrained to have at most 10 solutions, and redundant solutions are discarded with the crowding-distance-based downsizing mechanism from Section~\ref{sec:downsize}. Notice that as more iterations are run, not only does $T_e$ approximate $F_e$ closer at solution points, but the points are also more evenly distributed across the entire front. In Figure~\ref{fig:dynamic_approx}, $T_e$ does not have a size constraint. In this case $T_e$ almost exactly matches $F_e$.

\subsection{3-objective Sphere and Rastrin Domains}
\label{sec:3obj_experiments}
\begin{figure}[h]
    \centering
    \includegraphics[width=\columnwidth]{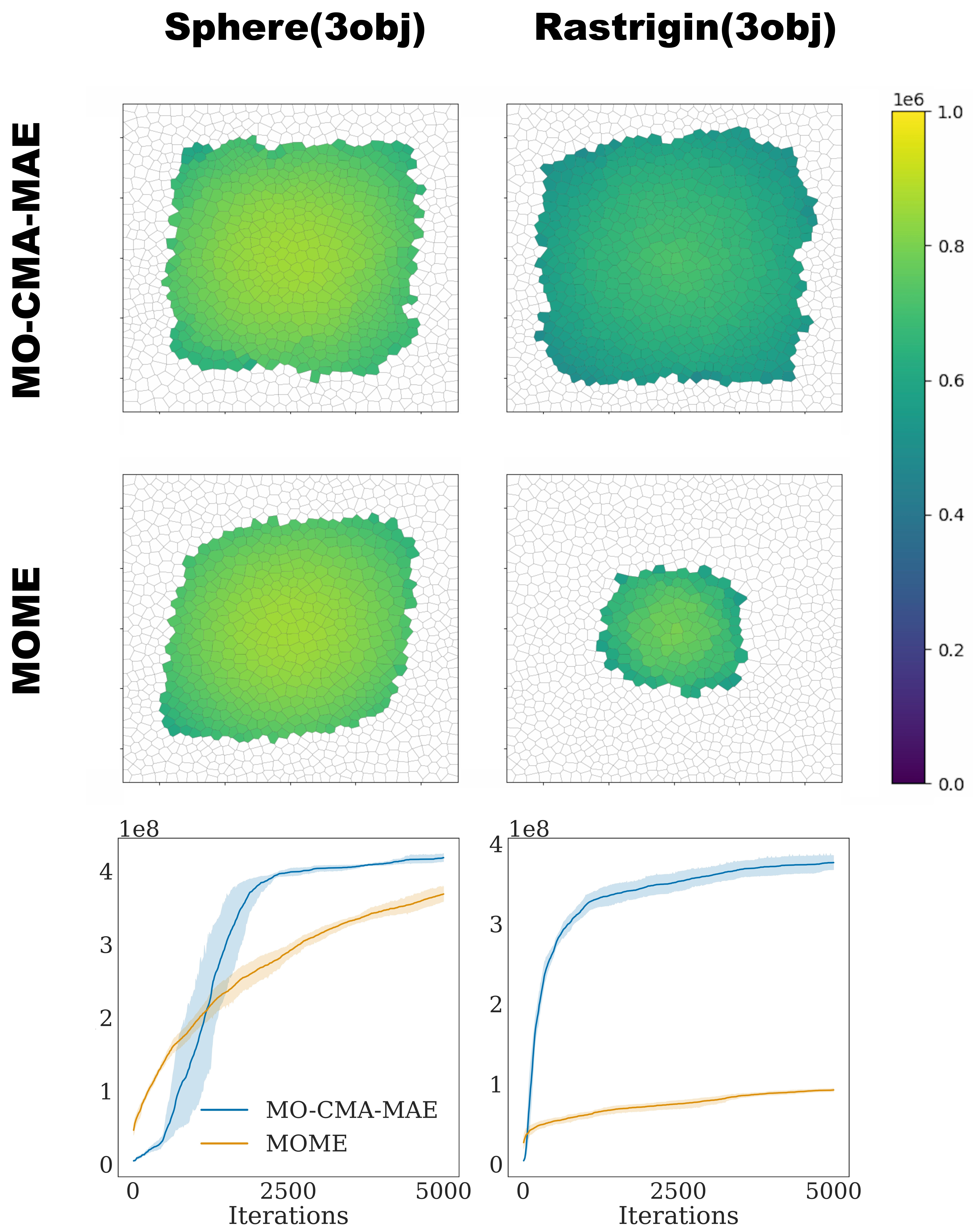}
    \caption{Archive heatmaps and MOQD-scores obtained by MO-CMA-MAE and MOME on 3-objective sphere and rastrigin domains.}
    \label{fig:3obj}
\end{figure}

To demonstrate the performance of MO-CMA-MAE on 3-objective optimization domains, we modify the 2-objective sphere and rastrin domains, defined in Equation~\ref{eq:sphere} and Equation~\ref{eq:rastrigin} respectively, to have 3 shift values $\lambda_1 = 4$, $\lambda_2 = 0$, and $\lambda_3 = -4$. We show in Figure~\ref{fig:3obj} the performances of MO-CMA-MAE and MOME on these two domains. Both algorithms use the same set of hyperparameters as in the main experiment.

% \subsection{Bisection Error Tolerance $\epsilon$}

\subsection{Static vs. Dynamic Archive}
\label{sec:staticvdynamic}
\begin{figure}[h]
    \centering
    \includegraphics[width=0.7\columnwidth]{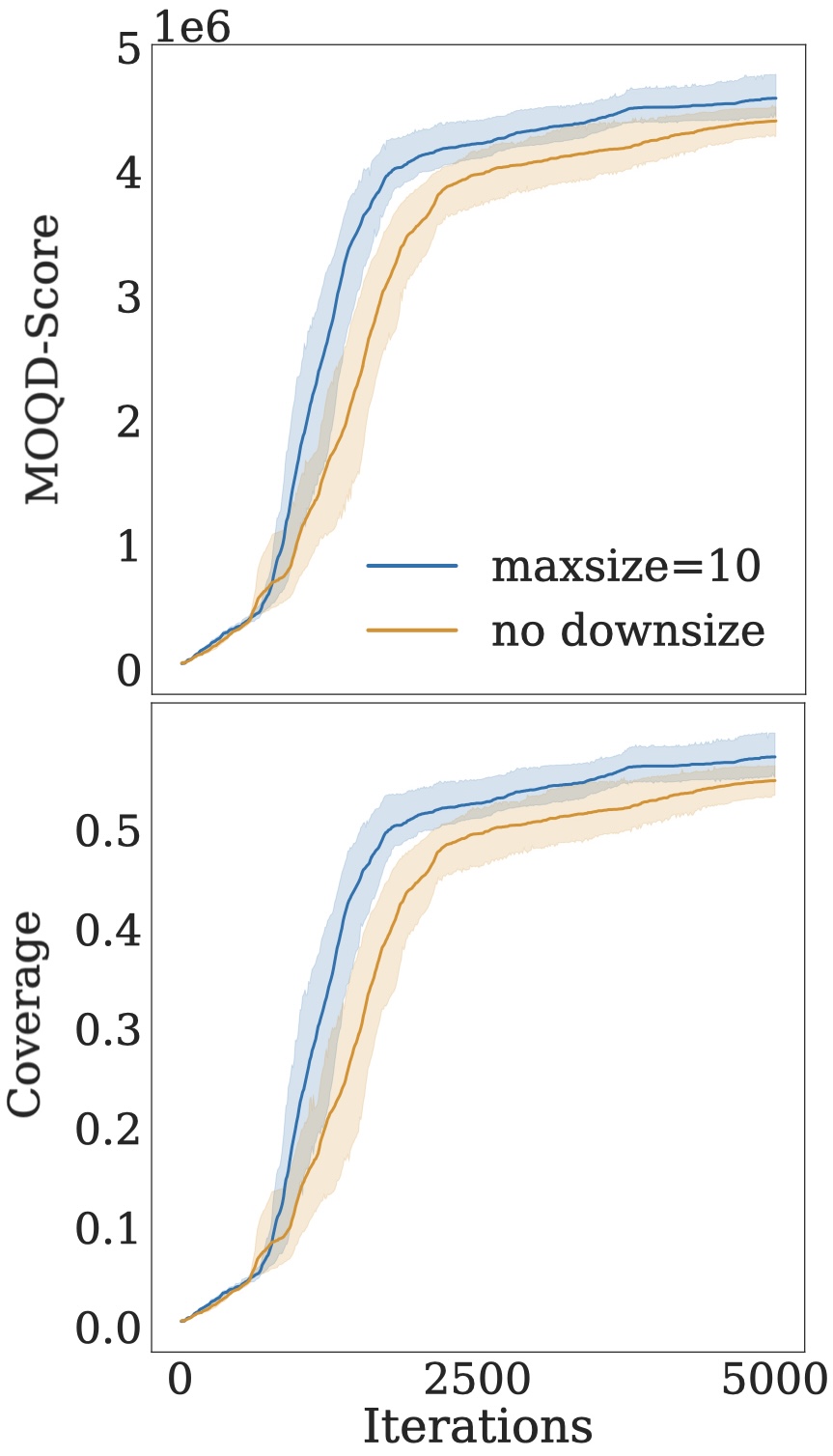}
    \caption{MOQD-scores and archive coverages achieved by static and dynamic versions of MO-CMA-MAE. The static version restricts maximum threshold front size to $10$ and uses \textit{cycle} restart. The dynamic version has no restriction on maximum threshold front size and uses \textit{basic} restart.}
    \label{fig:static_vs_dynamic}
\end{figure}

The static archive version of MO-CMA-MAE features a \textit{cycle} restart rule, which prevents MO-CMA-MAE from getting trapped in a loop of repeatedly adding and dropping the same solution. However, since this \textit{cycle} restart rule restarts the search at a new location in measure space, it also introduces an additional measure space exploration mechanism on top of our HVI-based CMA-ES search. In order to investigate how MO-CMA-MAE performs in the absence of archive size limitation and \textit{cycle} restart, and also to validate that the exploration caused by \textit{cycle} restart is not the main driving force behind MO-CMA-MAE, we compare the performances of static and dynamic archive MO-CMA-MAE on the sphere domain. Both versions use the same set of hyperparameters as in the main experiment, except the dynamic archive version does not limit the maximum threshold front size, and uses the \textit{basic} restart rule for CMA-ES. As shown in Figure~\ref{fig:static_vs_dynamic}, although the dynamic archive version has slightly worse performance, the difference is not significant.

\subsection{Pareto Front visualization for the Sphere domain}
\begin{figure}[h]
    \centering
    \includegraphics[width=0.8\columnwidth]{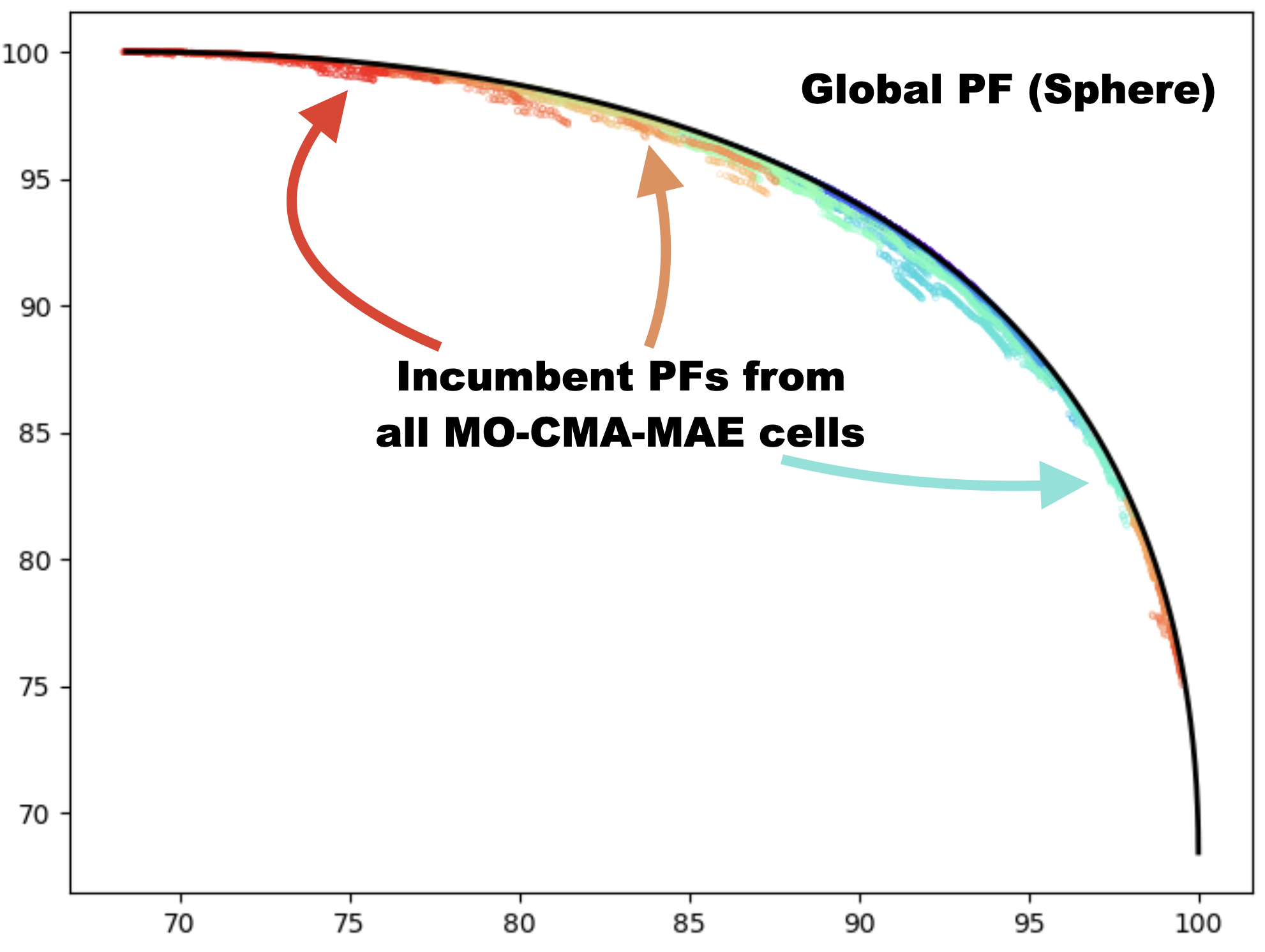}
    \caption{The incumbent local PFs from all cells found by MO-CMA-MAE collectively approximate the global PF in the \textit{sphere} domain.}
    \label{fig:local_v_global_PF}
\end{figure}

As an MOQD algorithm, MO-CMA-MAE is primarily designed to find a set of local Pareto Fronts rather than a single global Pareto Front. However, as briefly shown in Section~\ref{sec:main_results}, MO-CMA-MAE was able to approximate both ends of the global Pareto Front in the Sphere domain, thanks to its utilization of adaptive CMA-ES search. Here we further illustrate this behavior by plotting the incumbent Pareto Fronts from all cells found by MO-CMA-MAE alongside the ground truth global Pareto Front in the \textit{sphere} domain. As shown in Figure~\ref{fig:local_v_global_PF}, the incumbent local PFs from all cells collectively approximate the global PF in this domain.

\begin{figure*}[h]
    \centering
    \includegraphics[width=\linewidth]{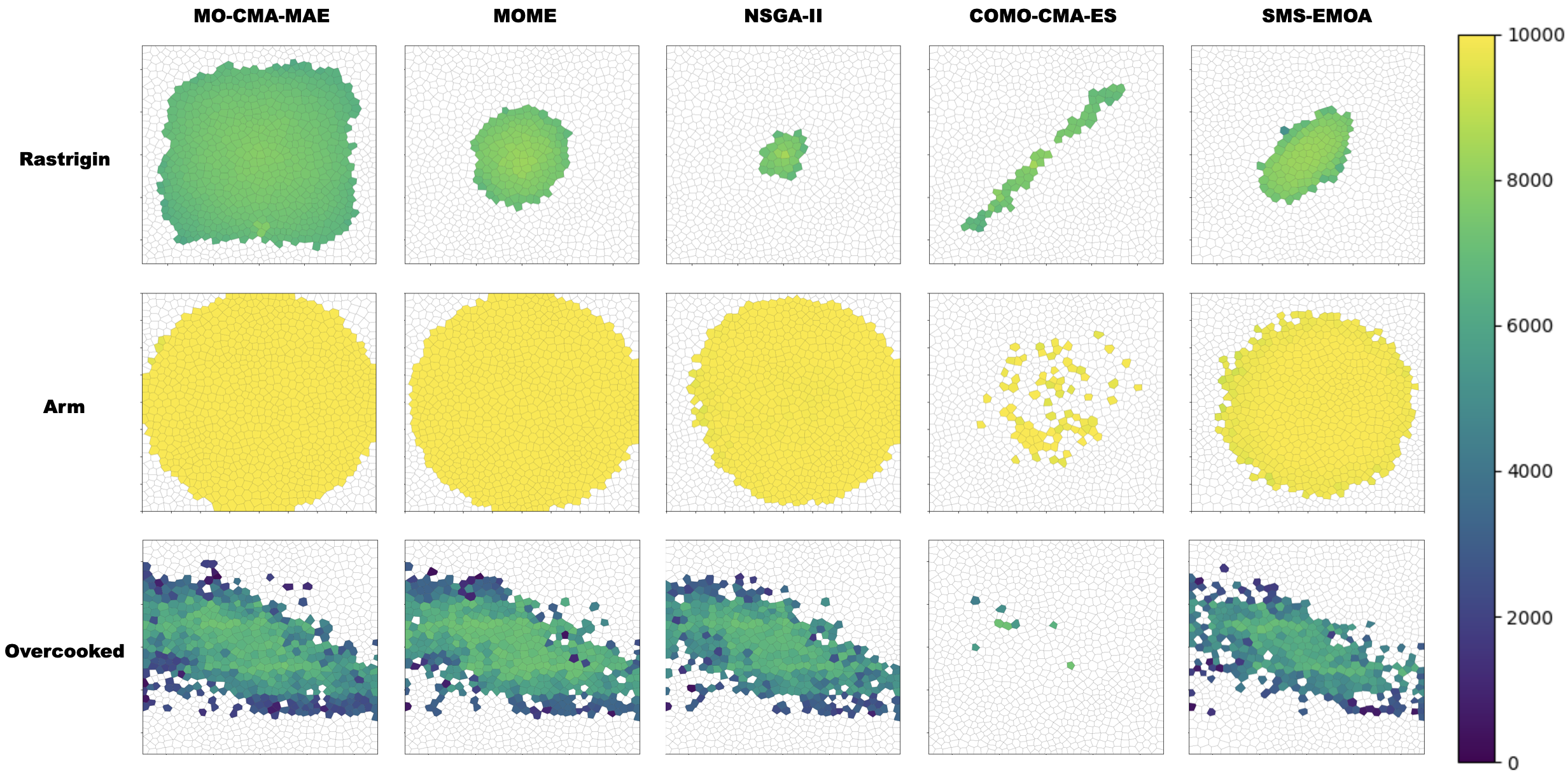}
    \caption{Heatmaps representing the passive MOQD archives for the \textit{rastrigin}, \textit{arm}, and \textit{overcooked} domains.}
    \label{fig:other_akv}
\end{figure*}

\subsection{Passive MOQD Archives for Rastrigin, Arm, and Overcooked Domains}
\label{sec:other_akv}
We show in Figure~\ref{fig:other_akv} the passive MOQD archives for the \textit{rastrigin}, \textit{arm}, and \textit{overcooked} domains after 5000 iterations of each algorithm. Consistent with observations from the \textit{sphere} domain (Figure~\ref{fig:sphere_akv}), MO-CMA-MAE and MOME demonstrate superior archive coverage compared to NSGA-II, SMS-EMOA, and COMO-CMA-ES.

\section{Statistical Test Results}
\label{sec:statistical_test_results}
We include the one-way ANOVA results on MOQD-scores across all domains in Table~\ref{tab:one_way_anova_results}, and the pairwise Tukey's HSD test comparisons of MOQD-scores in Table~\ref{tab:tukey_results1} and \ref{tab:tukey_results2}.

\begin{table}[h]
\caption{One-way ANOVA results on MOQD-scores across all domains.}
\label{tab:one_way_anova_results}\centering\resizebox{\columnwidth}{!}{\begin{tabular}{lcccc}
\toprule
 & \multicolumn{2}{c}{MOQD-Score} & \multicolumn{2}{c}{Coverage} \\
\cmidrule(lr){2-3} \cmidrule(lr){4-5}
 & F-value & p-value & F-value & p-value \\
\midrule
Sphere     &   F(4, 20) = 978.88  & 1.26E-22 &   F(4, 20) = 900.56  & 2.88E-22 \\
Rastrigin  &   F(4, 20) = 1087.19 & 4.43E-23 &   F(4, 20) = 1142.18 & 2.71E-23 \\
Arm        &   F(4, 20) = 331.79  & 5.64E-18 &   F(4, 20) = 343.00  &  4.07E-18 \\
Overcooked &   F(4, 20) = 1018.13 & 4.60E-18 &   F(4, 20) = 1490.04 & 2.67E-19 \\
\bottomrule
\end{tabular}
}
\end{table}

\clearpage

\begin{table*}[h]
\caption{Pairwise MOQD-score comparisons in the \textit{sphere} and \textit{rastrigin} domains.}
\label{tab:tukey_results1}\centering\resizebox{1.0\linewidth}{!}{\begin{tabular}{lrrrrrrrrrr}
\toprule
 & \multicolumn{5}{c}{Sphere} & \multicolumn{5}{c}{Rastrigin} \\
\midrule
  & MO-CMA-MAE & MOME & NSGA-II & COMO-CMA-ES & SMS-EMOA & MO-CMA-MAE & MOME & NSGA-II & COMO-CMA-ES & SMS-EMOA \\ 
  \cmidrule(lr){2-6} \cmidrule(lr){7-11}
MO-CMA-MAE & N/A & $>$ & $>$ & $>$ & $>$ & N/A & $>$ & $>$ & $>$ & $>$ \\
MOME & $<$ & N/A & $>$ & $>$ & $>$ & $<$ & N/A & $>$ & $>$ & $>$ \\
NSGA-II & $<$ & $<$ & N/A & $>$ & $-$ & $<$ & $<$ & N/A & $-$ & $<$ \\
COMO-CMA-ES & $<$ & $<$ & $<$ & N/A & $-$ & $<$ & $<$ & $-$ & N/A & $<$ \\
SMS-EMOA & $<$ & $<$ & $-$ & $-$ & N/A & $<$ & $<$ & $>$ & $>$ & N/A \\
\bottomrule
\end{tabular}
}
\end{table*}

\begin{table*}[h]
\caption{Pairwise MOQD-score comparisons in the \textit{arm} and \textit{overcooked} domains.}
\label{tab:tukey_results2}\centering\resizebox{1.0\linewidth}{!}{\begin{tabular}{lrrrrrrrrrr}
\toprule
 & \multicolumn{5}{c}{Arm} & \multicolumn{5}{c}{Overcooked} \\
\midrule
  & MO-CMA-MAE & MOME & NSGA-II & COMO-CMA-ES & SMS-EMOA & MO-CMA-MAE & MOME & NSGA-II & COMO-CMA-ES & SMS-EMOA \\ 
  \cmidrule(lr){2-6} \cmidrule(lr){7-11}
MO-CMA-MAE & N/A & $-$ & $>$ & $>$ & $>$ & N/A & $-$ & $>$ & $>$ & $>$ \\
MOME & $-$ & N/A & $>$ & $>$ & $>$ & $-$ & N/A & $>$ & $>$ & $>$ \\
NSGA-II & $<$ & $<$ & N/A & $>$ & $-$ & $<$ & $<$ & N/A & $>$ & $>$ \\
COMO-CMA-ES & $<$ & $<$ & $<$ & N/A & $<$ & $<$ & $<$ & $<$ & N/A & $<$ \\
SMS-EMOA & $<$ & $<$ & $-$ & $>$ & N/A & $<$ & $<$ & $<$ & $>$ & N/A \\
\bottomrule
\end{tabular}
}
\end{table*}

\end{document}